%% file: iclr2027_conference.tex
\documentclass[11pt]{article}

\usepackage[margin=1in]{geometry}

\input{math_commands.tex}

\usepackage{url}
\usepackage{graphicx}
\usepackage[numbers,sort&compress]{natbib}
\usepackage{caption}
\usepackage{algorithm}
\usepackage{algorithmic}
\usepackage{amsmath}
\usepackage{amssymb}
\usepackage{wrapfig}
\usepackage{booktabs}
\usepackage{array}
\usepackage{comment}

\usepackage{newfloat}
\usepackage{listings}
\usepackage[hidelinks]{hyperref}

\title{Beyond Masks and Trajectories: Flow-Guided Latent Action Injection for Stable Surgical Video Generation}
\date{}

\author{%
\begin{tabular}{@{}>{\centering\arraybackslash}p{0.31\textwidth}>{\centering\arraybackslash}p{0.31\textwidth}>{\centering\arraybackslash}p{0.31\textwidth}@{}}
Tsz-Yui Qin\textsuperscript{*} &
Siyu Zhou\textsuperscript{*} &
Chi-Keung Tang \\
\href{mailto:tyqin@connect.ust.hk}{\small\texttt{tyqin@connect.ust.hk}} &
\href{mailto:szhoubx@connect.ust.hk}{\small\texttt{szhoubx@connect.ust.hk}} &
\href{mailto:cktang@cs.ust.hk}{\small\texttt{cktang@cs.ust.hk}} \\[0.8em]
\multicolumn{3}{c}{The Hong Kong University of Science and Technology}
\end{tabular}
}

\newcommand{\degname}[1]{\texttt{#1}}

\begin{document}

\maketitle
\vspace{-2.2em}
\begingroup
\renewcommand{\thefootnote}{}
\footnotetext{\textsuperscript{*}Equal contribution.}
\endgroup

\begin{abstract}
Surgical video generation holds substantial potential for surgical education, simulation, and data augmentation, yet generating surgical videos with realistic and clinically plausible motion remains challenging. Most existing methods rely on auxiliary conditions, such as masks, trajectories, depth, or reference videos, to achieve visually plausible synthesis. Yet, these auxiliary conditions typically require additional manual annotation or specialized acquisition, making it difficult to scale such methods beyond small, curated datasets. This motivates the need for a reference-free architecture capable of generating high-quality surgical video without requiring auxiliary visual conditions at inference time. We propose \textbf{FLAIR}, a \textbf{F}low-guided \textbf{L}atent \textbf{A}ction \textbf{I}njection framework for \textbf{R}eference-free surgical video generation. FLAIR learns action priors from optical flow of real surgical videos, dynamically predicts corresponding latent action representation from an input prompt, and injects it into a frozen base model to generate surgical videos with improved action consistency. We further construct \textbf{SurgActionClip-30K}, the first large-scale surgical vision dataset comprising action-centric segmented clips and structured caption labels, addressing the persistent lack of fine-grained, action-centric surgical datasets. Lastly, we introduce \textbf{SurgMetrics}, the first surgical domain-specific evaluation metrics for quantifying the quality of generated surgical videos, addressing the persistent absence of clinically grounded evaluation standards in this domain. Extensive experiments demonstrate that FLAIR enables generating high-quality surgical videos using text-only inference without auxiliary conditions, and validation in SurgMetrics demonstrates its strength in alignment with human perception compared to traditional metrics. 
\end{abstract}

\section{Introduction}

Surgical videos are widely used in medical education, skill assessment, and the development of downstream surgical AI systems \citep{loukas2018video, eckhoff2023sages, king2025use}. Unlike general-domain videos, they contain procedure-constrained actions and fine-grained instrument-tissue interactions in which small motion deviations can
alter clinical meaning \citep{nwoye2022rendezvous}. 
Existing recordings are mostly long and untrimmed, lacking action-centric clips for learning motion patterns.


Recent surgical video generation methods mainly adapt general-domain generative models to surgical scenes through domain fine-tuning, text-guided diffusion, or additional structural conditioning \citep{li2024endora,cho2024surgen,li2025ophora,chen2025surgsora,biagini2025hierasurg,rapuri2026saw}. While these approaches have demonstrated the feasibility of surgical video synthesis and achieved improved visual realism and controllability, their controllability is often achieved through explicit inference-time conditions, including reference frames, segmentation masks, depth maps, or tool trajectories \citep{yeganeh2024visage,sivakumar2025sg2vid,chen2025surgsora,biagini2025hierasurg,rapuri2026saw}, which are costly to obtain and unavailable in many low-resource educational or clinical settings \citep{yang2025see}. This raises the need for reference-free, text-only motion guidance.

Short action-centric clips make this problem tractable. Relative to the broad motion space of general-domain videos, surgical videos occupy a substantially narrower region constrained by anatomy, instrument affordances, and procedural workflow. Within this domain, instances of a specific action concentrate further into a compact action-conditioned region, allowing text alone to select a plausible motion mode without predicting arbitrary frame-level dynamics. Text therefore only needs to identify the intended action and map it to this restricted region, while the action's canonical execution paradigm can be learned from recurring examples in real surgical videos.

Motivated by the observation that optical flow naturally deemphasizes static background while highlighting the motion of surgical interactions in videos, we propose \textbf{FLAIR}, a \textbf{F}low-guided \textbf{L}atent \textbf{A}ction \textbf{I}njection framework for \textbf{R}eference-free surgical video generation. This framework introduces motion priors learned from real surgical videos into a text-to-video diffusion model, allowing text prompts to guide surgical video generation without requiring auxiliary inference-time conditions. Specifically, FLAIR first learns compact surgical action embeddings from optical flow, then trains a text-to-action predictor to map surgical prompts into the learned latent action space. The predicted action embedding is injected into a frozen diffusion Transformer through a lightweight adapter, providing motion guidance during generation without relying on masks, trajectories, depth maps, or reference videos. Experiments show that FLAIR improves motion stability and action consistency over domain-adapted baselines, demonstrating that text prompts can guide not only the appearance but also the motion of surgical video generation.

In summary, this work makes the following contributions: (1) We propose \textbf{FLAIR}, a framework that learns flow-derived latent action priors during training and enables reference-free, text-only inference without auxiliary conditions. (2) We construct \textbf{SurgActionClip-30K}, the first large-scale surgical video dataset with action-centric segmentation and template-guided captions.  (3) We introduce \textbf{SurgMetrics}, the first objective evaluation metrics tailored for surgical video generation, which provide domain-specific measurements with six artifact-specific scores and an independent domain-typicality score. (4) Extensive experiments across standard metrics and SurgMetrics demonstrate that FLAIR consistently improves motion stability, action consistency, and surgical-domain plausibility over existing baselines, supporting the feasibility of text-only surgical video generation.

\section{Related Work}
\subsection{Surgical Vision Datasets}

Existing surgical vision datasets can be broadly categorized into two paradigms. The first consists of raw, procedure-level datasets such as Cholec80~\citep{twinanda2016endonet}, SurgicalActions160~\citep{SurgicalActions160}, and AutoLaparo~\citep{wang2022autolaparo}, which provide long, untrimmed recordings. These datasets' unsegmented nature renders them ill-suited for downstream generative training. A second paradigm addresses this limitation by segmenting and re-organizing raw surgical recordings into shorter, concentrated samples. Surg-396K~\citep{wang2025endochat} aggregates three sources into instruction-style image-text pairs, while SurgLaVi~\citep{perez2026surglavi} curates nearly 240K narrative clip-caption pairs and organizes them hierarchically by task and step. Similar datasets, including SurgAtlas~\citep{bellos2026surgatlas}, SurgVLM-DB~\citep{zeng2506surgvlm}, and Surg-QA~\citep{li2024llava},  further extend this direction by scaling up data and enriching captions. Yet, they are still either uncaptioned or segmented in ways that fail to align with discrete action boundaries, leaving none with the precision and specificity required for fine-grained, action-centric downstream training. SurgActionClip-30K addresses this gap by consolidating diverse surgical sources into a unified collection of action-centric clips with standardized textual descriptions.

\subsection{Surgical Video Generation Models}
Recent surgical video generation methods have explored a wide range of architectures and conditioning strategies in pursuit of realistic surgical video synthesis. Endora \citep{li2024endora} and Ophora \citep{li2025ophora} establish foundations of video generation in endoscopic and ophthalmic settings, respectively. SurgSora \citep{chen2025surgsora} leverages initial-frame, RGB-D, and flow cues to strengthen motion and semantic control, while HieraSurg \citep{biagini2025hierasurg} uses segmentation maps and hierarchical procedure annotations to impose phase and action structure. SAW \citep{rapuri2026saw} further combines language with reference scenes, affordance masks, and tool-tip trajectories for controllable surgical rollouts. Yet, most existing methods rely on auxiliary conditions for controllability,  and acquiring such conditions at scale is often costly or impractical, fundamentally limiting scalability. Approaching the problem differently, FLAIR encodes optical-flow features as training-time supervision, enabling a reference-free, text-only inference architecture that generates realistic surgical videos without any auxiliary input during inference, offering a substantially more practical and scalable alternative to existing methods.

\subsection{Evaluation Metrics for Surgical Videos}
Surgical video generation evaluation remains largely underexplored. Some existing surgical video benchmarks focus on downstream discriminative tasks, such as phase recognition, workflow analysis, and action recognition, where metrics including accuracy, F1 score, and Jaccard index are used to validate generation performance \citep{twinanda2016endonet,nwoye2022rendezvous,wang2022autolaparo}. These metrics, however, are fundamentally unsuited for quantifying the quality of generated surgical videos. Therefore, current surgical video generation studies further adopt general-purpose video generation metrics, including FID, FVD, PSNR, Temporal IoU (TI), and Area Flicker (AF) \citep{heusel2017gans,unterthiner2018towards}. While these metrics provide useful measures of overall visual similarity or semantic correspondence, they are insensitive to surgical-specific errors such as implausible instrument-tissue interactions, instrument deformation, or domain shift. Recent efforts such as SurgVeo introduce expert-based evaluation protocols for surgical video generation, highlighting the gap between visual plausibility and surgical correctness \citep{chen2025far}. SurgMetrics addresses this gap by introducing a multi-axis, surgical scenario-specific evaluator, with six dedicated heads each quantifying a distinct dimension of surgical video quality. These per-axis scores together form a standardized, surgery-aware metric for evaluating generated surgical videos.

\section{Dataset Preparation}

\subsection{Cross-domain data sources}
We assemble temporally compact surgical action clips from multiple domains into a unified corpus, \textbf{SurgActionClip-30K}, covering diverse anatomy, procedures, and actions. Guided by the cross-specialty action taxonomy of BSA-10~\citep{xu2026generalized}, we carefully select 11 datasets, including CholecT50~\citep{nwoye2022rendezvous}, AutoLaparo~\citep{wang2022autolaparo}, etc., to improve coverage of surgical actions and their procedural contexts. The resulting corpus includes laparoscopic, endoscopic, and microsurgical procedures, covering representative actions such as dissection, coagulation, clipping, retraction, and aspiration. All sources are converted into a unified clip-metadata-caption schema; source-level details are provided in Appendix~\ref{sec:supp_data_sources}.

\subsection{Action-centric clip extraction and captioning}
We segment action-centric surgical clips from the original long videos using available phase, step, action, or temporal-interval annotations. Consecutive segments with consistent procedural labels are merged, while non-operative and uninformative segments are removed. For datasets without explicit instrument annotations, we employ a single-class YOLO detector \citep{yolo11_ultralytics} to estimate instrument presence and split clips when visible tool configurations change. Then, each resulting clip is paired with a structured caption following an instrument-action-target representation. We use Hulu-Med-4B \citep{jiang2025hulu} to generate initial descriptions, which are constrained by available annotations and verified for consistency with clip metadata. Beyond the core action triplet,  we also enrich each caption with other detailed surgical annotations, including the procedure, anatomical context, tissue response, and imaging conditions to provide semantic supervision for learning surgical motion priors. Further details and examples of the filtering and captioning procedures are provided in Appendices~\ref{sec:supp_clip_extraction} and~\ref{sec:supp_captioning}.

\section{Methods}
\subsection{Framework Overview}
\begin{figure*}[t]
    \centering    \includegraphics[width=0.95\textwidth]{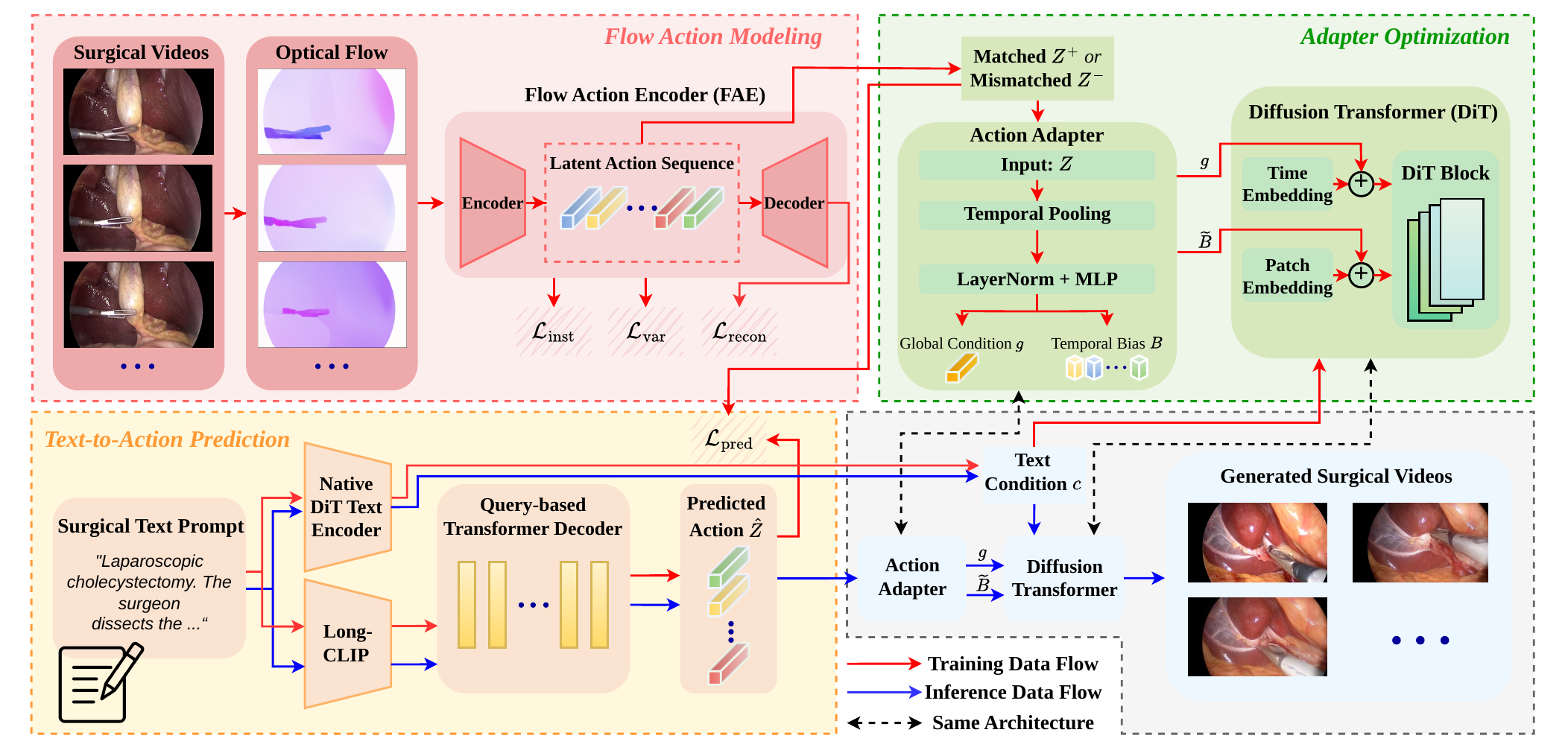}
    \caption{
    {\bf Overview of FLAIR.} During training, the FAE extracts flow-derived latent action sequences $Z$, which supervise text-to-action prediction and adapter optimization. At inference, the FAE is removed, and the surgical prompt provides both the predicted motion prior $\widehat{Z}$ and the backbone-native text condition, enabling reference-free surgical video generation.
    }\vspace{-0.2in}\label{fig:flair_overview}
\end{figure*}


Figure~\ref{fig:flair_overview} illustrates FLAIR. The Flow Action Encoder (FAE) first learns flow-derived latent action sequences $Z$ from real surgical videos. The text-to-action predictor estimates the corresponding motion prior $\widehat{Z}$ from a prompt, while the action adapter injects global and temporal conditions into a frozen, domain-adapted diffusion transformer. At inference, the FAE is removed and generation requires only surgical prompt.

\subsection{Flow Action Modeling}
\label{sec:latent_action}

To learn a compact representation of surgical motion, we train FAE on our \textbf{SurgActionClip-30K} dataset. Given a real surgical clip $V=\{I_t\}_{t=1}^{T}$, we first estimate the adjacent-frame optical-flow sequence $F=\{F_t\}_{t=1}^{T-1}$ using RAFT \citep{teed2020raft}. Compared with RGB inputs, optical flow suppresses static appearance and emphasizes motion arising from instrument and tissue interactions. The FAE, denoted by $E_{\phi}$, independently maps each optical-flow representation to an $\ell_2$-normalized latent action token:
\begin{equation}
\mathbf{z}_k=E_{\phi}(F_k),\qquad Z=[\mathbf{z}_1,\ldots,\mathbf{z}_K],\qquad \mathbf{z}_k\in\mathbb{R}^{d_z},
\label{eq:flow_action_encoding}
\end{equation}
where $F_k$ denotes the optical-flow representation between frames $I_k$ and $I_{k+1}$, $\mathbf{z}_k\in\mathbb{R}^{d_z}$ is its encoded latent action token, and $Z$ is the ordered latent sequence formed by all $K=T-1$ frame transitions. For an 81-frame clip, we obtain $K=80$ ordered motion tokens with $d_z=128$. Their temporal organization follows the order of the corresponding frame transitions, while each token is encoded independently. A lightweight decoder $D_{\varphi}$ reconstructs the input flow representation as $\widehat{F}_k=D_{\varphi}(\mathbf{z}_k)$, encouraging the latent space to preserve motion information without introducing RGB appearance cues.

The FAE is optimized using reconstruction and anti-collapse objectives:
\begin{equation}
\mathcal{L}_{\mathrm{FAE}}=\mathcal{L}_{\mathrm{recon}}+\lambda_{\mathrm{inst}}\mathcal{L}_{\mathrm{inst}}+\lambda_{\mathrm{var}}\mathcal{L}_{\mathrm{var}}.
\label{eq:fae_objective}
\end{equation}
The reconstruction loss $\mathcal{L}_{\mathrm{recon}}$ preserves the information required to recover the observed motion representation. The instance-discrimination objective $\mathcal{L}_{\mathrm{inst}}$ keeps distinct flow observations distinguishable in the latent space, while the variance-floor regularizer $\mathcal{L}_{\mathrm{var}}$ prevents dimensional collapse without encouraging unbounded latent variation. Detailed formulations and optimization settings are provided in Appendices~\ref{app:objectives} and~\ref{sec:supp_fae_setup}.


\subsection{Action Adapter and Optimization}
\label{sec:action_injection}

Given a flow-derived latent action sequence $Z\in\mathbb{R}^{K\times d_z}$, we introduce a lightweight action adapter $A_{\omega}$ that converts the motion prior into global and temporally varying conditions for a frozen, domain-adapted diffusion Transformer. During training, the adapter receives latent sequences extracted by the frozen FAE, whereas at inference they are replaced by the text-predicted sequence $\widehat{Z}$.

The adapter first reduces the temporal resolution of the input sequence and maps the pooled tokens into a shared feature space:
\begin{equation}
\begin{aligned}
\overline{Z}
&=\operatorname{Pool}_{K\rightarrow L}(Z),\\
H
&=\operatorname{MLP}\left(\operatorname{LN}(\overline{Z})\right),\\
\mathbf{g}
&=A_{\mathrm{global}}(H),\\
B=&A_{\mathrm{temp}}(H)=[\mathbf{b}_1,\ldots,\mathbf{b}_L].
\end{aligned}
\label{eq:action_adapter}
\end{equation}
Here, $\operatorname{Pool}_{K\rightarrow L}$ partitions the $K$ ordered input tokens into $L$ consecutive groups and mean-pools each group, producing $\overline{Z}\in\mathbb{R}^{L\times d_z}$. The operators $\operatorname{LN}$ and $\operatorname{MLP}$ denote layer normalization and token-wise projection through a multi-layer perceptron, respectively, getting $H$ as the resulting adapter feature sequence. Furthermore, the global head $A_{\mathrm{global}}$ aggregates $H$ into a clip-level condition $\mathbf{g}\in\mathbb{R}^{d_h}$, while the temporal head $A_{\mathrm{temp}}$ produces $B\in\mathbb{R}^{L\times d_h}$, where $\mathbf{b}_{\ell}$ is the condition at temporal position $\ell$ and $d_h$ is the hidden dimension of the video backbone.

Secondly, in case of any difference in latent length, we resize $B$ to obtain $\widetilde{B}=[\widetilde{\mathbf{b}}_1,\ldots,\widetilde{\mathbf{b}}_{L_v}]$, where $L_v$ is the temporal length of the video latent. Ultimately, the global and temporal conditions are injected additively:
\begin{equation}
\widetilde{\mathbf{e}}_u
=
\mathbf{e}_u+s\mathbf{g},
\qquad
\widetilde{\mathbf{x}}_{\ell,n}
=
\mathbf{x}_{\ell,n}+s\widetilde{\mathbf{b}}_{\ell},
\label{eq:action_injection}
\end{equation}
where $\mathbf{e}_u$ is the diffusion-time embedding at denoising time $u$, $\mathbf{x}_{\ell,n}$ is the patch embedding at temporal index $\ell$ and spatial index $n$, and $s$ controls the conditioning strength. The temporal condition $\widetilde{\mathbf{b}}_{\ell}$ is shared across spatial positions at the same latent time, allowing the subsequent Transformer blocks to process a shared time-specific motion context.

We optimize the adapter through the conditioned forward pass of the frozen generator using its native flow-matching objective. To prevent the adapter from ignoring the supplied motion sequence, we compare the matched sequence $Z^{+}$ from the target video with a mismatched sequence $Z^{-}$ from another sample:
\begin{equation}
\mathcal{L}_{\mathrm{rank}}=
\text{max}(0, m+\mathcal{L}_{\mathrm{den}}(Z^{+})
-\mathcal{L}_{\mathrm{den}}(Z^{-})),
\label{eq:adapter_ranking}
\end{equation}
where $\mathcal{L}_{\mathrm{den}}$ is the backbone-native flow-matching loss, and $m$ is the ranking margin.
Finally, the overall objective is
\begin{equation}
\mathcal{L}_{\mathrm{adapter}}
=
\mathcal{L}_{\mathrm{den}}(Z^{+})
+
\lambda_{\mathrm{rank}}\mathcal{L}_{\mathrm{rank}}
+
\mathcal{L}_{\mathrm{reg}}.
\label{eq:adapter_objective}
\end{equation}
The ranking term encourages the matched motion prior to provide more useful conditioning than an unrelated sequence, while $\mathcal{L}_{\mathrm{reg}}$ denotes auxiliary adapter regularization. Detailed formulations of $\mathcal{L}_{\mathrm{den}}$ and $\mathcal{L}_{\mathrm{reg}}$, together with optimization settings, are provided in Appendices~\ref{app:objectives} and~\ref{sec:supp_adapter_setup}.

\begin{figure*}[t]
\centering
\includegraphics[width=\textwidth]{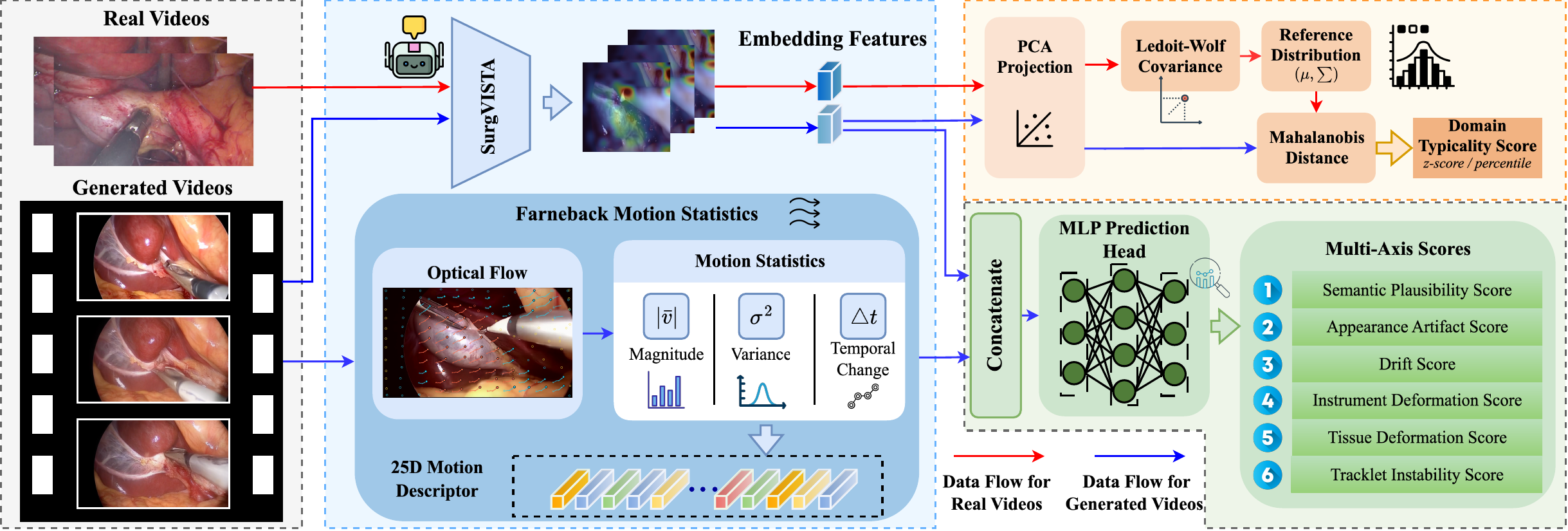}
\caption{
{\bf Overview of SurgMetrics.} During inference, SurgMetrics extracts SurgViSTA representation and Farneback statistics from input videos to predict six artifact scores: ``semantic implausibility, appearance artifact, drift, instrument deformation, tissue deformation, and tracklet instability''. Then, SurgMetrics compute the Domain Distance of input video independently by comparing its SurgViSTA features with a real surgical reference distribution to compute how much a video belongs to the surgical domain. SurgMetrics incurs low computational overhead at inference time.
}
\label{fig:surgmetrics}
\end{figure*}

\subsection{Text-to-Action Prediction}
\label{sec:text_action}

The FAE requires real-video optical flow and is therefore unavailable at inference. We instead train a text-to-action predictor to estimate a latent motion prior from the surgical prompt.

To encode the detailed descriptions in our structured surgical captions, we use a frozen Long-CLIP text encoder $E_{\mathrm{text}}$ \citep{zhang2024long}. Given a surgical prompt $p$, the encoder produces
\begin{equation}
\mathbf{h}_p=E_{\mathrm{text}}(p),\qquad \mathbf{h}_p\in\mathbb{R}^{d_t},
\end{equation}
where $\mathbf{h}_p$ is the pooled text feature with the dimension of $d_t$. Long-CLIP accommodates our detailed surgical captions and is used only for latent-action prediction; the diffusion backbone retains its separate native text encoder.

We introduce $K$ learnable queries $Q=[\mathbf{q}_1,\ldots,\mathbf{q}_K]$, where $\mathbf{q}_k$ represents the $k$-th latent position. A query-based Transformer decoder predicts all tokens in parallel:
\begin{equation}
\widehat{Z}=P_{\psi}(\mathbf{h}_p,Q)=[\widehat{\mathbf{z}}_1,\ldots,\widehat{\mathbf{z}}_K]\in\mathbb{R}^{K\times d_z},
\label{eq:text_action_prediction}
\end{equation}
where $P_{\psi}$ is the predictor and $\widehat{\mathbf{z}}_k\in\mathbb{R}^{d_z}$ is the predicted token at position $k$.

Together, with the FAE and Long-CLIP frozen, the predictor is trained by
\begin{equation}
\mathcal{L}_{\mathrm{pred}}=
\frac{1}{K}\sum_{k=1}^{K}\left[1-\cos\left(\widehat{\mathbf{z}}_k,\mathbf{z}_k\right)\right]
+\lambda_z\frac{1}{K}\sum_{k=1}^{K}\left\|\widehat{\mathbf{z}}_k-\mathbf{z}_k\right\|_2^2,
\label{eq:text_action_objective}
\end{equation}
where $\mathbf{z}_k$ is the FAE target token and $\lambda_z$ balances cosine alignment and Euclidean regression. Since one prompt may admit multiple valid executions, $\widehat{Z}$ is treated as a text-conditioned motion prior rather than an exact trajectory. At inference, $\widehat{Z}$ replaces $Z$ as the adapter input, transferring motion knowledge learned from real surgical videos to reference-free, text-only generation while preserving the backbone's native semantic conditioning.

\section{SurgMetrics: Surgical Video Evaluation}
\label{sec:surgmetrics}
Figure~\ref{fig:surgmetrics} provides an overview of SurgMetrics. Given a generated surgical video, SurgMetrics extracts visual features and motion statistics, fuses them into a shared representation, and outputs six surgical artifact scores. In parallel, an independent domain-distance branch measures the deviation of the generated clip from real laparoscopic videos, providing an estimate of its domain typicality.

\subsection{Lightweight feature fusion and multi-axis scoring}
Given a video clip $V$, SurgMetrics extracts two complementary feature groups. A fine-tuned SurgViSTA encoder \citep{yang2026large} produces a representation $\phi_{\mathrm{sv}}(V)$ that captures surgical semantics, appearance, and coarse temporal context. Meanwhile, Farneback optical flow statistics \citep{farneback2003two} produce a compact motion descriptor $\phi_{\mathrm{mot}}(V)$ characterizing motion magnitude, variation, and temporal irregularity. The two feature groups are independently projected
before fusion:
\begin{equation}
\mathbf{f}(V) = \left[ P_{\mathrm{sv}}\!\left(\phi_{\mathrm{sv}}(V)\right), P_{\mathrm{mot}}\!\left(\phi_{\mathrm{mot}}(V)\right) \right].
\label{eq:surgmetrics_features}
\end{equation}
We first fine-tune the final blocks of SurgViSTA on controlled surgical artifacts and subsequently freeze the encoder when training the lightweight fusion network.

A shared prediction network maps $\mathbf{f}(V)$ to six artifact scores:
\begin{equation}
\widehat{\mathbf{y}} = h(\mathbf{f}(V)) = [ \hat{y}_{\mathrm{sem}}, \hat{y}_{\mathrm{app}}, \hat{y}_{\mathrm{drift}}, \hat{y}_{\mathrm{inst}}, \hat{y}_{\mathrm{tissue}}, \hat{y}_{\mathrm{trk}} ].
\label{eq:surgmetrics_outputs}
\end{equation}
The six axes measure evidence of Semantic Implausibility, Appearance Artifacts, Temporal or Global-motion Drift, Instrument Deformation, Tissue Deformation, and Local Tracklet Instability. The appearance axis includes both spatial degradation and temporally inconsistent appearance effects such as brightness flicker and excessive motion blur. Higher values indicate stronger evidence of the corresponding artifact.

\begin{table}[t]
\caption{
{\bf Comparison of LoRA-FLAIR} on three base models with baselines and related works on general video metrics and SurgMetrics. LoRA has already achieved a greater level of TI and AF owing to the great preparation of our training datasets. LoRA-FLAIR further surpasses the result of LoRA on both general metrics and SurgMetrics.
``\textbf{--}'' denotes data not comparable or not available. ``V.CLIP'' denotes VideoCLIP \citep{xu2021videoclip}. 
}
\centering
\scriptsize
\setlength{\tabcolsep}{2pt}
\renewcommand{\arraystretch}{1.05}

\resizebox{\textwidth}{!}{%
\begin{tabular}{@{}l|ccccc|ccccccc@{}}
\toprule
\textbf{Method}
&\multicolumn{5}{c|}{\textbf{General Video Metrics}}
&\multicolumn{7}{c}{\textbf{SurgMetrics}}\\

\cmidrule(lr){2-6}
\cmidrule(l){7-13}

&\textbf{V.CLIP}$\uparrow$
&\textbf{FID}$\downarrow$
&\textbf{FVD}$\downarrow$
&\textbf{TI}$\uparrow$
&\textbf{AF}$\downarrow$
&\textbf{Sem}$\downarrow$
&\textbf{App}$\downarrow$
&\textbf{Drift}$\downarrow$
&\textbf{Inst}$\downarrow$
&\textbf{Tissue}$\downarrow$
&\textbf{Track}$\downarrow$
&\textbf{Dom}$\downarrow$\\

\midrule

Cosmos-H-Surgical
&\textbf{--}&\textbf{--}&\textbf{--}&0.780&8.52
&\textbf{--}&\textbf{--}&\textbf{--}&\textbf{--}&\textbf{--}&\textbf{--}&\textbf{--}\\

KVLR
&\textbf{--}&\textbf{--}&\textbf{--}&0.893&4.48
&\textbf{--}&\textbf{--}&\textbf{--}&\textbf{--}&\textbf{--}&\textbf{--}&\textbf{--}\\

\midrule

Wan2.1 Base
&8.80&246.8&3466&0.814&1.50
&\textbf{--}&\textbf{--}&\textbf{--}&\textbf{--}&\textbf{--}&\textbf{--}&4.228\\

Wan2.1 LoRA
&8.82&126.3&1340&0.877&0.11
&0.097&\textbf{0.132}&0.027&0.059&0.085&0.100&0.931\\

\textbf{Wan2.1 LoRA-FLAIR (ours)}
&\textbf{9.08}&\textbf{117.2}&\textbf{1148}&\textbf{0.913}&\textbf{0.03}
&\textbf{0.029}&0.175&\textbf{-0.005}&\textbf{0.027}
&\textbf{-0.023}&\textbf{-0.001}&1.165\\

\midrule

Hunyuan-1.5 Base
&7.66&209.1&2655&0.833&3.57
&--&--&--&--&--&--&4.373\\

Hunyuan-1.5 LoRA
&\textbf{8.41}&151.9&2533&0.857&0.32
&0.028&0.182&0.024&0.054&0.238&0.193&3.388\\

\textbf{Hunyuan-1.5 LoRA-FLAIR (ours)}
&8.11&\textbf{122.9}&\textbf{1367}&\textbf{0.914}&\textbf{0.05}
&\textbf{0.005}&\textbf{0.024}&\textbf{-0.013}&\textbf{0.019}
&\textbf{0.072}&\textbf{0.039}&1.771\\

\midrule

CogVideoX-2B Base
&7.11&216.3&3227&0.826&6.88
&--&--&--&--&--&--&5.231\\

CogVideoX-2B LoRA
&\textbf{8.47}&219.0&1772&0.944&3.94
&\textbf{0.081}&0.525&0.009
&0.063&0.168&0.120&3.446\\

\textbf{CogVideoX-2B LoRA-FLAIR (ours)}
&8.27&\textbf{202.3}&\textbf{1592}&\textbf{0.964}&\textbf{0.83}
&0.164&\textbf{0.487}&\textbf{0.001}&\textbf{0.061}&\textbf{0.094}
&\textbf{0.070}&3.527\\

\bottomrule
\end{tabular}%
}
\label{tab:all_metrics}
\vspace{-0.2in}
\end{table}

\subsection{Axis-specific artifact supervision}
SurgMetrics is trained using 21 controlled degradation families designed specifically for surgical settings, which are applied to clean laparoscopic clips. These degradations cover temporal and global-motion errors, appearance corruption, semantic discontinuities, instrument failures, and tissue failures. The complete degradation definitions are provided in Appendix~\ref{sec:supp_degradations}.

Each degradation may contribute to one or more related artifact axes. Given a degradation recipe $\mathcal{D}$, the target for axis $k$ is composed using a noisy-OR rule:
\begin{equation}
y_k = 1- \prod_{d\in\mathcal{D}} \left( 1-w_{k,d}s_d \right),
\label{eq:surgmetrics_noisyor}
\end{equation}
where $s_d\in[0,1]$ is the severity of degradation $d$, and $w_{k,d}$ specifies its contribution to axis $k$. This formulation allows multiple artifact signals to accumulate while keeping the target within $[0,1]$. Clean clips are assigned zero-valued targets on all axes.

Because the six artifact types occur with different frequencies, the prediction heads are optimized with an axis-balanced Smooth L1 objective:
\begin{equation}
\mathcal{L}_{\mathrm{artifact}}=\frac{1}{N} \sum_{n=1}^{N}\sum_{k=1}^{6}\alpha_k\,\operatorname{SmoothL1}\left(\hat{y}_{n,k}, y_{n,k} \right),
\label{eq:surgmetrics_loss}
\end{equation}
where $\alpha_k$ is determined from the positive prevalence of axis $k$ in the training set. Training follows a curriculum that first emphasizes clean and single-degradation examples before introducing clips containing multiple simultaneous artifacts. Full degradation definitions and label-routing weights are provided in Appendix~\ref{sec:supp_degradations}, while network dimensions and training details are provided in Appendix~\ref{sec:supp_training_setup}.

\subsection{Surgical domain distance}
Artifact severity does not by itself determine whether a video belongs to the target surgical domain. Thus, we add an extra distance score that models domain membership in SurgMetrics. SurgViSTA features extracted from clean real training videos are projected using principal component analysis (PCA) \citep{abdi2010principal}, after which a Ledoit--Wolf shrinkage covariance model \citep{ledoit2004well} is fitted in the projected space. For an input video $V$, the domain distance is
\begin{equation}
d_{\mathrm{dom}}(V)=\sqrt{\left(\mathbf{z}(V)-\boldsymbol{\mu}\right)^{\top}\boldsymbol{\Sigma}^{-1}\left(\mathbf{z}(V)-\boldsymbol{\mu}\right)},
\label{eq:surgmetrics_domain}
\end{equation}
where $\mathbf{z}(V)$ is the PCA-projected SurgViSTA feature, and $\boldsymbol{\mu}$ and $\boldsymbol{\Sigma}$ are estimated exclusively from clean real surgical training videos. We report the resulting domain z-score as the domain-typicality measure. Values near zero indicate consistency with the real surgical reference distribution, whereas large positive values indicate stronger out-of-domain deviation. This separation is important because a video may exhibit few visible artifacts while remaining out-of-domain, or may belong to the laparoscopic domain while containing severe generation errors.

\section{Experiments}

\subsection{Quantitative and Qualitative Results}

\begin{figure*}[t]
    \centering   
    \vspace{-0.2in}\includegraphics[width=\textwidth]{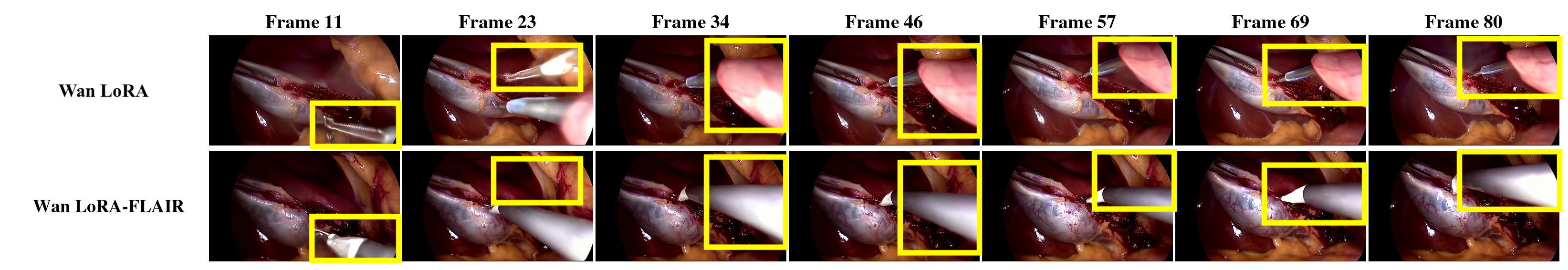}
    \caption{
    {\bf Qualitative temporal comparison} between Wan2.1 LoRA and Wan2.1 LoRA-FLAIR motion conditioning, with key changes highlighted by bounding boxes. Under identical generation settings, Wan2.1 LoRA-FLAIR produces a more coherent motion trajectory and more sustained instrument–tissue interaction across the generated surgical video. Seven synchronized frames are shown from frame 0 to frame 80.
    }
    \label{fig:temporal_comparison}
    \vspace{-0.2in}
\end{figure*}

\paragraph{Inference Prompt Set.}
As no standardized surgical benchmark currently exists, we construct a 200-prompt benchmark based on the BSA-10 taxonomy \citep{xu2026generalized}, covering ten basic surgical actions across eight procedures. The first 180 prompts form a balanced grid of six procedures, ten actions, and three phrasing variants, while the remaining 20 extend the evaluation to VATS lobectomy and hepatectomy for external generalization. Each prompt follows a structured instrument--action--target formulation, with fine-grained vocabulary grounded in available surgical annotations and manually extended to procedures not covered by those annotations. We use the same prompt set for all experimental comparisons; its construction and full distribution are detailed in Appendix~\ref{sec:supp_prompt_set}.

\textbf{Baselines and Comparisons.}
We evaluate FLAIR across three base models, Wan2.1-T2V-1.3B \citep{wan2025wan}, HunyuanVideo-1.5 \citep{wu2025hunyuanvideo}, and CogVideoX-2B \citep{yang2025cogvideox}, against two baselines: a Base-model-only baseline and a LoRA baseline fully fine-tuned on our dataset until convergence, serving as weak and strong points of comparison, respectively. However, many existing surgical video generation methods do not release model checkpoints or generated outputs, which prevents our re-evaluation with SurgMetrics. We therefore further consider Cosmos-H-Surgical \citep{he2025surgworld} and KVLR \citep{li2026articulated}, but omit FID/FVD for these methods because their reference-conditioned inference settings differ from our reference-free setting, and omit CLIP/SurgMetrics scores where raw generated videos are unavailable.


\textbf{Evaluation of General Metrics.}
As shown in Table~\ref{tab:all_metrics}, general-purpose metrics including VideoCLIP, TI and AF, show that FLAIR maintains competitive visual quality and correct execution specified by the prompt while introducing higher action consistency. However, these metrics evaluate mainly overall distribution similarity and temporal consistency, but cannot directly measure whether the generated motions are surgically plausible or ensure human alignment on surgical aspects.

\textbf{Evaluation on SurgMetrics.}
We further evaluate the generated videos with SurgMetrics, which provides surgical-specific diagnosis through artifact axes. Compared with conventional domain fine-tuned baselines, FLAIR achieves lower artifact evidence across the majority of axes, particularly in motion-related factors such as Drift, Instrument Deformation, and Tracklet Instability. This indicates that learning action-level motion priors from optical-flow-derived representations improves procedural motion consistency beyond appearance-level adaptation. FLAIR demonstrates better execution of instrument–action–target interaction specified by the prompt.

\paragraph{Out of Domain.}
We exclude videos with domain distance above 4 from SurgMetrics comparisons because they fall outside the calibrated distribution, while values below approximately 2 are treated as in-domain. Reference values are provided in Appendix~\ref{sec:supp_domain_distance}.



\textbf{Qualitative Results: Human Perception.}
Figure~\ref{fig:temporal_comparison} compares Wan2.1 LoRA and Wan2.1 LoRA-FLAIR under identical generation settings across frames $t \in [0-81]$. LoRA-only exhibits inconsistent instrument motion that escalates into structural artifacts, including a spurious second instrument and a persistent geometric fusion between instrument and tissue in later frames. In contrast, Wan2.1 LoRA-FLAIR remains stable throughout, producing a more coherent motion trajectory and sustained instrument–tissue interaction that closely aligns with real surgical dynamics. Additional qualitative results are provided in Appendix~\ref{sec:supp_more_qualitative}.

\subsection{Human-Alignment Validation for SurgMetrics}

We conducted a human evaluation of 90 generated laparoscopic videos, with ten videos from each of nine model configurations. Four specialist surgeons with approximately 7--21 years of clinical experience independently rated each video on a five-point ordinal scale. Videos were anonymized and randomly shuffled, with only the corresponding prompts shown. Evaluators assessed visible defects corresponding to the six SurgMetrics dimensions, as well as overall video quality and the correctness of the ``instrument--action--target`` interaction specified by the prompt.

To evaluate whether SurgMetrics captures native generation failures beyond its synthetic training degradations, we compared each axis with the corresponding human rating. Spearman correlations ranged from 0.746 to 0.867, indicating strong agreement with human judgments. Inter-rater reliability was also high, with ICC(2,4) values of 0.721--0.832 across artifact axes and 0.742--0.825 for action correctness \citep{koo2016guideline}. After adjusting for model identity, the correlation with overall human-rated quality remained strong, decreasing from 0.845 to 0.731, indicating that the association was not solely driven by differences among models. FLAIR also received lower interaction-error ratings than the LoRA baseline across all three backbones, with paired confidence intervals excluding zero. Overall, these results show strong human alignment of SurgMetrics and consistent improvements in action correctness with FLAIR. The current validation is restricted to laparoscopic videos; full validation details are provided in Appendix~\ref{sec:supp_human_alignment}.

\subsection{Ablation Study on FAE Motion Representation}
 We evaluate whether the latent representation learned by the Flow Action Encoder (FAE) contains real motion information, by testing whether it is usable beyond optical-flow reconstruction. We freeze the FAE and train a one-step world-model probe to predict the next surgical frame:
  \begin{equation}
      \hat{\mathbf{x}}_{t+1}
      =
      \mathbf{x}_{t}
      +
      g_{\theta}(\mathbf{x}_{t}, \mathbf{z}_{t}),
    \end{equation}
    \begin{equation}
      \mathbf{z}_{t}
      =
      \mathrm{FAE}(\mathrm{Flow}(\mathbf{x}_{t},\mathbf{x}_{t+1})),
  \end{equation}
where $g_{\theta}$ is a $128\times128$ residual U-Net. The frozen 128-dimensional motion latent $\mathbf{z}_{t}$ is injected into the U-Net bottleneck through FiLM modulation. The probe predicts the frame residual.

\begin{wraptable}{r}{0.5\linewidth}
\vspace{-0.2in}
    \caption{
    {\bf Next-frame prediction error} under different FAE latent conditions.
    Percentages denote relative MSE changes compared with the no-latent baseline.
    }
    \vspace{-0.1in}\label{tab:fae_world_model_ablation}
    \centering
    \small
    \setlength{\tabcolsep}{9pt}

    \begin{tabular}{lc}
        \toprule
        \textbf{Latent Condition} &
        \textbf{Next-Frame MSE} $\downarrow$ \\
        \midrule
        No latent
        & 0.002521 (reference) \\
        Zero latent
        & 0.002525 $(+0.15\%)$ \\
        Shuffled FAE latent
        & 0.002601 $(+3.17\%)$ \\
        FLAIR FAE latent
        & \textbf{0.002356} $\mathbf{(-6.55\%)}$ \\
        \bottomrule
    \end{tabular}
    \vspace{-0.3in}
\end{wraptable}    

Table~\ref{tab:fae_world_model_ablation} shows that FAE latent of FLAIR reduces next-frame MSE by more than 6.55\% relative to the no-latent, zero-latent, and shuffled-latent controls. These results demonstrate that the FAE latent preserves transition-specific motion information that can be exploited by an independently trained predictor. FAE training and checkpoint-selection details are provided in Appendix~\ref{sec:supp_fae_setup}.

\section{Conclusion}

We introduce FLAIR, a reference-free framework that encodes motion priors from optical flow into a latent action space, dynamically injecting them into a frozen video diffusion model to generate surgical videos with improved motion stability and action consistency without auxiliary conditions at inference time. To support this direction, we further contribute SurgActionClip-30K and SurgMetrics, together establishing an integrated foundation spanning modeling, data, and evaluation for scalable, action-aware surgical video generation. We hope these resources catalyze future research at the intersection of generative modeling and surgical video understanding.

\section*{AI Use Statement}

Generative AI tools were used in two parts of this work. 
First, Hulu-Med-4B was used during dataset construction to generate initial structured captions for surgical video clips, conditioned on available procedural and annotation metadata. These captions were subsequently constrained and checked for consistency with the corresponding clip metadata, as described in the paper. 
Second, a large language model was used to assist with language editing and improving the clarity of the manuscript. 
All methodological design, experiments, data processing, analysis, interpretation of results, and scientific conclusions were carried out and verified by the authors.

\section*{Ethics Statement}

This work uses anonymized surgical video data obtained from publicly available research datasets, and the authors had no access to personally identifiable patient information. SurgActionClip-30K is planned for public release, including metadata and processed video clips where redistribution is permitted by the corresponding source licenses. The human-alignment study involved four surgeons with specialist qualifications and 7--21 years of clinical experience; all participated with informed consent and received no financial compensation. Hulu-Med-4B was run locally for caption generation, and no surgical video data were transmitted to external services. FLAIR-generated videos are intended solely for research and education and should not be used for clinical decision-making, diagnosis, treatment planning, or intraoperative guidance.

\section*{Reproducibility Statement}

We provide detailed descriptions of the data construction pipeline, model architecture, training objectives, optimization procedure, evaluation protocol, and ablation settings in the main paper and Appendices~\ref{app:dataset_construction}--\ref{app:method_details}.
Appendix~\ref{app:dataset_construction} specifies the construction and quality-control procedures for SurgActionClip-30K; Appendix~\ref{app:objectives} gives the complete objectives used to train the flow action encoder, text-to-action predictor, and action adapter; and Appendix~\ref{app:method_details} reports the training and inference configurations used for the evaluated video backbones.
We also provide the construction procedure and full distribution of the 200-prompt evaluation benchmark, and release the complete prompt set to facilitate consistent cross-model evaluation. 
Additional details on SurgMetrics, including its controlled degradation families, supervision design, validation protocol, and ablation experiments, are provided in Appendices~\ref{sec:supp_degradations}, \ref{sec:surgmetrics_validation}, and~\ref{sec:supp_ablation_surgmetrics}.

To further support reproducibility, SurgActionClip-30K and its associated metadata are intended to be released subject to the redistribution and licensing constraints of the original source datasets.

\bibliography{iclr2027_conference}
\bibliographystyle{plainnat}

\newpage
\appendix
\section{Formal Motivation}
\label{app:formal_motivation}

This section of the appendix provides more discussion on the main design choices underlying FLAIR and SurgMetrics.

\subsection{Why the Surgical Domain Differs Fundamentally from General Video Domains}

Surgical video differs from general-domain video along two fundamental axes: semantic granularity and visual dynamics~\citep{loukas2018video}. These differences explain why many architectures and training strategies that succeed in general-domain video generation do not transfer directly to the surgical setting.

\paragraph{Semantic granularity.}
In general-domain video, an action is typically recognizable at a coarse level: whether a person is walking, jumping, or waving conveys the intended semantics even under substantial variation in speed, trajectory, or viewpoint~\citep{simonyan2014two,wang2016temporal,carreira2017quo}. Surgical actions do not enjoy this tolerance. A small deviation in instrument angle, contact depth, or approach trajectory can change the clinical meaning of a motion entirely, e.g., the difference between grasping and gripping too forcefully, or between coagulation and inadvertent tissue damage, is often a matter of a few pixels of instrument-tissue overlap rather than a qualitatively different motion pattern~\citep{nwoye2022rendezvous}. This means that surgical video generation must be evaluated and supervised at a substantially finer level of motion granularity than general-domain video, where coarse trajectory-level similarity is often sufficient. Architectures designed to capture broad, semantically distinct motion categories in general-domain video are therefore not naturally suited to distinguishing these fine-grained, clinically meaningful variations, motivating our use of optical flow, rather than coarse action labels or trajectory sketches, as a supervisory signal for learning surgical motion priors.

\paragraph{Visual dynamics and background stability.}
General-domain videos are also naturally characterized by large motion magnitudes and substantial scene change: subjects move across the frame, cameras pan or cut, and backgrounds shift as the scene evolves. Surgical videos exhibit the opposite tendency. In a typical laparoscopic or endoscopic clip, the majority of the visual field, the surrounding anatomy, the camera framing, and the overall scene layout, remains largely static, with only the instrument and the immediately interacting tissue undergoing subtle, localized motion. This asymmetry between a mostly static background and a small, semantically critical region of motion is precisely what motivates our use of optical flow as the supervisory signal in FLAIR: optical flow naturally suppresses the dominant, uninformative static background and isolates exactly the small-magnitude, instrument-tissue motion that carries surgical meaning~\citep{ng2018actionflownet,sun2018optical}, whereas RGB-based or trajectory-sketch-based conditioning, which are more effective when scene-level motion is itself informative, provide comparatively little additional signal in a domain where the background barely changes.

Taken together, these two properties, high semantic sensitivity to small motion deviations and a naturally low-motion, background-static visual regime, distinguish surgical video from the general-domain videos that most existing video generation architectures and conditioning strategies are designed around~\citep{hu2023videocontrolnet,liang2024movideo,koroglu2025onlyflow}, and directly motivate FLAIR's core design choice of encoding motion priors from optical flow.

\subsection{Why FLAIR Helps}

Building on this domain-level distinction, the effectiveness of FLAIR is further grounded in an observation about the structure of surgical motion itself. Relative to the broad motion space of general-domain videos, surgical videos occupy a substantially narrower region constrained by anatomy, instrument affordances, and procedural workflow: instruments can only move in ways that are mechanically feasible within the surgical field, and their trajectories are further shaped by the anatomical target and the stage of the procedure. Within this already-narrow domain, instances of a specific action concentrate further into a compact, action-conditioned region, since surgeons executing the same basic action (e.g., clipping, dissection, retraction) tend to follow similar approach angles, instrument-tissue contact patterns, and motion durations across patients and procedures. This structure means that, unlike general-domain text-to-video generation, where a prompt must implicitly resolve an enormous space of plausible motions, a surgical prompt only needs to identify the intended action and route it into this compact region; the canonical execution pattern within that region can then be learned directly from recurring examples in real surgical videos. FLAIR is designed to exploit exactly this property: the flow action encoder distills the compact, action-conditioned motion patterns present in real surgical clips into a latent space, and the text-to-action predictor learns to map a surgical prompt into this same compact region rather than predicting arbitrary frame-level dynamics from scratch. This explains why injecting a flow-derived latent action prior, rather than relying on the base diffusion model's general-purpose motion prior, yields more stable and action-consistent surgical motion at inference time.

\subsection{Why SurgMetrics is Helpful}

SurgMetrics is motivated by a complementary observation about the nature of failure modes in generated surgical video. Unlike open-ended, general-domain video generation, where failure modes can be highly diverse and difficult to anticipate, the ways in which current surgical video generation models fail are comparatively enumerable: instruments intermittently disappearing or reappearing, geometric drift and structural instability across frames, implausible instrument or tissue deformation, and physically inconsistent instrument-tissue interaction recur across models and backbones~\citep{cho2024surgen,li2024endora,yeganeh2024visage,chen2025surgsora,chen2025far}, as also observed qualitatively in our comparisons (Appendix~\ref{sec:supp_more_qualitative}). This enumerability suggests that, rather than requiring costly expert annotation of every generated sample, a substantial fraction of surgical-domain failure modes can instead be synthesized by deliberately applying controlled degradations to clean, real laparoscopic clips. SurgMetrics is built on this premise: by constructing 21 degradation families that emulate these recurring artifact types and training a multi-axis evaluator to detect them, we obtain a scalable and reproducible surrogate for surgical-domain video quality assessment, without depending on general-purpose metrics~\citep{heusel2017gans,unterthiner2019fvd,huang2024vbench} that are insensitive to surgical-specific errors or on expert review for every comparison. Our human-alignment validation (Appendix~\ref{sec:supp_human_alignment}) further supports the practical value of this approach, showing that scores derived from these synthesized artifacts correlate strongly with human judgments of real generated surgical videos.

\section{SurgActionClip-30K Dataset Construction}
\label{app:dataset_construction}
\begin{figure*}[t]
    \centering
    \includegraphics[width=\textwidth]{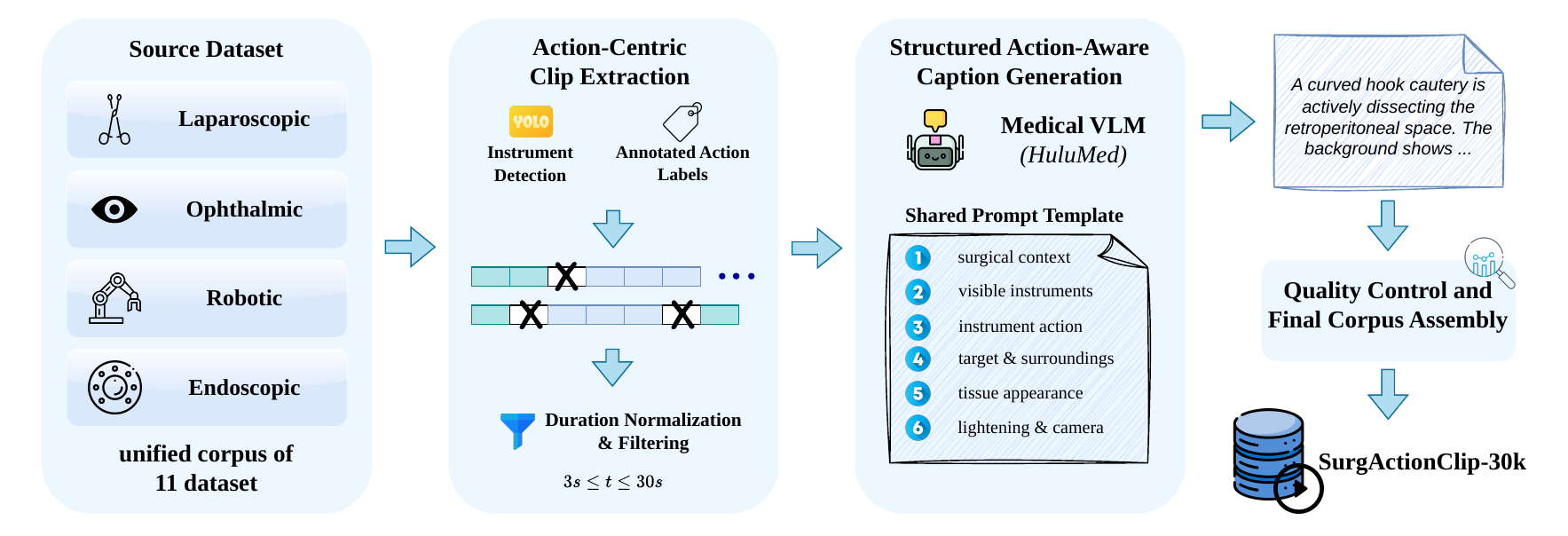}
    \caption{
    Overview of the SurgActionClip-30K construction pipeline. We first aggregate eleven surgical source datasets spanning laparoscopic, ophthalmic, robotic, and endoscopic domains. Action-centric clips are extracted using available temporal annotations, instrument-aware refinement, and duration-based filtering. Each clip is then paired with a structured action-aware caption generated by a medical VLM using a shared prompt template covering surgical context, instruments, actions, anatomical targets, tissue response, and imaging conditions. Finally, quality control and consistency checks are applied to assemble the unified SurgActionClip-30K corpus.
    }
    \label{fig:dataset}
\end{figure*}
This section provides additional details on the source selection, temporal segmentation, instrument-aware refinement, caption generation, and quality control used to construct SurgActionClip-30K. The overall pipeline is shown in Figure \ref{fig:dataset}, which aims to convert heterogeneous, procedure-level surgical videos into a common set of temporally compact clips, each describing a localized surgical activity and paired with action-aware textual supervision. We use the finest temporal annotation available in each source rather than assuming that all datasets provide action labels at the same granularity.

\subsection{Source Datasets and Unified Representation}
\label{sec:supp_data_sources}

The source datasets span laparoscopic, robotic, rigid-endoscopic, and ophthalmic procedures. Their native annotations range from coarse surgical phases to fine-grained steps, action classes, instrument identities, and explicit temporal intervals. This diversity broadens the coverage of anatomy, procedural context, instrument appearance, and instrument--tissue interaction in the resulting corpus.

Irrespective of the source format, every retained sample is converted to a common representation containing a unique clip identifier, its source case or video, the available procedural label, instrument information when available, the clip path, and a structured caption. Source-specific labels are preserved in the metadata instead of being forcibly mapped to a single closed vocabulary. This avoids conflating differently defined phases or actions while allowing a shared data loader to consume all sources.

\begin{table*}[t]
    \caption{Compact summary of the eleven source datasets, their temporal supervision, and the number of clips retained in SurgActionClip-30K.}
    \label{tab:surgactionclip_sources}
    \centering
    \footnotesize
    \setlength{\tabcolsep}{4pt}
    \renewcommand{\arraystretch}{1.15}
    \begin{tabular}{@{}p{0.15\textwidth}p{0.34\textwidth}p{0.33\textwidth}r@{}}
        \toprule
        \textbf{Source} & \textbf{Procedure or domain} & \textbf{Boundary supervision} & \textbf{Retained clips} \\
        \midrule
        CholecT50 & Laparoscopic cholecystectomy & Instrument--verb--target triplets & 3,773 \\
        M2CAI16 & Laparoscopic cholecystectomy & Phase labels; detector refinement & 3,086 \\
        AutoLaparo & Laparoscopic hysterectomy & Phase labels; detector refinement & 3,416 \\
        SLAM & Laparoscopic cholecystectomy & Native action groups & 221 \\
        PhaKIR & Laparoscopic surgery & Joint phase and instrument labels & 1,268 \\
        Surgical\-Actions160 & Laparoscopic action demonstrations & Native action classes & 75 \\
        HeiCo & Laparoscopic colorectal surgery & Phase labels; detector refinement & 4,411 \\
        OphNet & Ophthalmic microsurgery & Annotated temporal intervals & 1,661 \\
        SurgVU24 & Robotic surgery & Native clip and task boundaries & 700 \\
        PitVis & Endoscopic transsphenoidal surgery & Joint step and instrument labels & 6,445 \\
        ESD57 & Endoscopic submucosal dissection & Phase labels & 7,237 \\
        \midrule
        \textbf{Total} & & & \textbf{32,293} \\
        \bottomrule
    \end{tabular}
\end{table*}

\subsection{Action-Centric Clip Extraction}
\label{sec:supp_clip_extraction}

\paragraph{Annotation-driven candidate boundaries.} For datasets with frame-level or second-level phase, step, or action labels, we first run-length encode consecutive identical labels. Each maximal run becomes a candidate temporal interval. Datasets that already provide localized intervals, such as OphNet, use the annotated start and end times directly. For SLAM, consecutive short parts sharing the same patient, video, camera, and action identifiers are concatenated, while the annotated looping tail is removed before concatenation. CholecT50, SurgicalActions160, and SurgVU24 retain their native action-triplet, action-class, or clip boundaries, respectively.

\paragraph{Instrument-aware refinement.} When instrument annotations are available, they are preferred to automatically inferred labels. PhaKIR intervals are defined by stable \((\text{phase},\text{instrument set})\) pairs, and PitVis intervals are defined by stable \((\text{step},\text{instrument set})\) pairs. Consequently, a clip does not silently cross an annotated instrument change.

For laparoscopic sources without instrument annotations, we apply a single-class YOLO detector~\citep{yolo11_ultralytics} as a tool-presence model. The detector is evaluated at 5 frames per second with a confidence threshold of 0.25. Short detection gaps of at most 0.5 seconds are bridged to tolerate temporary occlusion and isolated false negatives. Longer tool-absent intervals create new boundaries, and sub-clips without a usable tool-present run are discarded. This refinement is restricted to visually compatible laparoscopic sources. It is not applied to ophthalmic microscopy or fisheye gastrointestinal endoscopy, where the domain gap would make detector-derived supervision unreliable.

\paragraph{Duration normalization and content filtering.} Except for short native action units in SLAM, candidate clips shorter than 3 seconds are removed. Long intervals are divided into non-overlapping pieces of at most 30 seconds without crossing the original annotation boundary. We also remove explicitly non-operative or uninformative labels. For example, the ``Others'' phase is excluded from ESD57, ``Undefined'' intervals are excluded from PhaKIR, and ``Non-functional,'' ``Step Interval,'' and ``Invalid Segment'' intervals are excluded from OphNet. Audio is removed when clips are written. These operations yield temporally compact samples while retaining the procedural label inherited from the source interval.

\subsection{Structured Action-Aware Caption Generation}
\label{sec:supp_captioning}

Each extracted clip is captioned with Hulu-Med-4B~\citep{jiang2025hulu} using a shared prompt template. The prompt supplies the available procedure, phase, step, action, and instrument metadata and asks the model to describe six semantic components: (1) procedural context, (2) visible instruments, (3) instrument action, (4) anatomical target and surrounding structures, (5) tissue appearance and response, and (6) lighting and camera perspective. The central semantic unit is an instrument--action--target description, while the remaining components provide contextual supervision useful for surgical video generation.

The instrument clause depends on the available supervision. If a source provides instrument identities, the prompt lists them as an exhaustive set and forbids additional instruments. If only detector-derived instrument counts are available, the prompt constrains the number of instruments mentioned but lets the model identify their types from the video. If neither identity nor a reliable count is available, the model uses open-vocabulary descriptions and is instructed to mention only instruments that are visible. Domain-appropriate instrument examples are supplied as guidance rather than as a closed label set. This tiered design reduces unsupported tool hallucination without treating cross-domain detector predictions as ground truth.

\paragraph{Caption examples.} For a CholecT50 clip annotated with the triplet \emph{grasper--grasp--gallbladder}, the generated caption states that the grasper holds and manipulates the gallbladder, then describes the surrounding liver and fatty tissue, the tissue surface, lighting, and camera stability. For a PitVis clip labeled \emph{nasal corridor creation}, the caption identifies the current operative context, describes the sphenoid-region anatomy and minor bleeding, and explicitly states that no instrument is visible. For an OphNet clip labeled \emph{corneal incision creation}, the caption identifies the keratome action, the corneal response, the iris and pupil in the background, and the close-up microscopic view. These examples illustrate how the same semantic template is instantiated across distinct surgical domains without forcing them into an identical vocabulary.

\paragraph{Prompt Engineering Template}

Each captioning prompt follows a fixed compositional template: \textit{procedure and surgical context} + \textit{visible instruments} + \textit{instrument--action--target interaction} + \textit{surrounding anatomy} + \textit{tissue appearance and response} + \textit{lighting and camera conditions}. Available procedure, phase, step, action, and instrument metadata are inserted as constraints, while Hulu-Med-4B is instructed to describe only information supported by the metadata or visible in the clip. When instrument identities are available, they are treated as an exhaustive set; when only instrument counts are available, the number of mentioned tools is constrained; otherwise, the model uses open-vocabulary descriptions restricted to clearly visible instruments.

\subsection{Quality Control and Final Corpus Assembly}
\label{sec:supp_quality_control}

We apply deterministic consistency checks after temporal extraction and captioning. Every clip must have a unique identifier, a readable video path, non-empty metadata, and a non-empty caption. The numbers of video files, metadata entries, and caption records are reconciled for every source. Captioning is checkpointed per clip and failed or empty generations are retried, allowing interrupted runs to resume without duplicating completed samples. The instrument constraints in the prompt are also derived from the same metadata record used to construct the clip, preventing cross-sample label mixing.

To prevent near-duplicate temporal content from leaking across partitions, the final training, validation, and test sets should be divided at the patient or source-video level before enumerating their clips. All clips originating from one case must remain in the same partition. The release manifest should report the final per-source retained counts and the case-level split used to obtain SurgActionClip-30K. SurgActionClip-30K will be made publicly available.

\section{Additional Details on FLAIR}
\label{app:flair_details}

\subsection{Inference Prompt Set}
\label{sec:supp_prompt_set}

To enable fair and reproducible cross-model comparison, we construct a 200-prompt benchmark. The construction of this prompt set draws on three complementary sources, described below.

\paragraph{Action and procedure taxonomy.}
The action and procedure skeleton is grounded in BSA-10~\citep{xu2026generalized}, which defines 10 Basic Surgical Actions (BSAs): \textit{Aspiration}, \textit{Clipping}, \textit{Coagulation}, \textit{Dissection}, \textit{Knot-tying}, \textit{Needle Grasping}, \textit{Needle Puncture}, \textit{Packaging}, \textit{Suture Pulling}, and \textit{Tissue Retraction}. These 10 actions are evaluated across 6 core surgical procedures (\textit{Cholecystectomy}, \textit{Gastrectomy}, \textit{Hysterectomy}, \textit{Intestinal Resection}, \textit{Nephrectomy}, \textit{Prostatectomy}) and 2 additional procedures reserved for external generalization testing (\textit{VATS Lobectomy}, \textit{Hepatectomy}).

\paragraph{"Instrument--verb--target" vocabulary.}
Fine-grained instrument--verb--target vocabulary is derived from CholecT50 triplet annotations in Cholec80~\citep{nwoye2022rendezvous}, comprising 5{,}209 clip-caption pairs. From these annotations we identify 90 unique triplets spanning 6 instrument classes, 9 verb classes, and 14 anatomical target classes (e.g., \textit{grasper} $\rightarrow$ \textit{grasp} $\rightarrow$ \textit{gallbladder}, \textit{clipper} $\rightarrow$ \textit{clip} $\rightarrow$ \textit{cystic-duct}, \textit{hook} $\rightarrow$ \textit{dissect} $\rightarrow$ \textit{peritoneum}, and \textit{irrigator} $\rightarrow$ \textit{aspirate} $\rightarrow$ \textit{fluid}.). As these triplets originate exclusively from Cholec80, they provide annotation-grounded coverage only for the cholecystectomy subset; instrument--action--target combinations for the remaining procedures (e.g., gastrectomy, hysterectomy, lobectomy) are manually extended based on standard surgical practice rather than dataset-derived triplet annotations.

\paragraph{Prompt engineering template.}
Each prompt follows a fixed compositional template: \textit{subject} + \textit{explicit BSA action} + \textit{instrument--verb--anatomical target} + \textit{surgical scene/background} + \textit{visual style} + \textit{aesthetic controls} + \textit{optional camera language}. For example: \textit{``A laparoscopic surgeon performs aspiration in a laparoscopic cholecystectomy field, using an irrigator to aspirate bile fluid pooling near the gallbladder bed after controlled perforation, tight close-up scope view, realistic surgical lighting, 4K.''} Two additional constraints are enforced across all prompts: (1) every prompt includes the fixed subject phrase \textit{``A laparoscopic surgeon''} to prevent video generation models from failing to recognize the endoscopic/surgical context; (2) every prompt explicitly states the BSA action label (e.g., \textit{``performs needle grasping''}) rather than leaving the action implicit in natural language.

\paragraph{Distribution of the 200 prompts.}
The first 180 prompts form a strict $6 \times 10 \times 3$ factorial grid (6 core procedures $\times$ 10 BSA actions $\times$ 3 phrasing variants per action), while the remaining 20 prompts extend coverage to two additional organs/procedures for external generalization testing and are not organized as a complete, balanced ten-action grid. Table~\ref{tab:prompt_set_distribution} summarizes the full distribution.

\begin{table}[t]
\caption{Distribution of the 200-prompt inference benchmark across procedures. The first 180 prompts (IDs 000--179) follow a strict 6-procedure $\times$ 10-action $\times$ 3-variant factorial design; the last 20 prompts (IDs 180--199) extend coverage to two external-validation procedures.}
\label{tab:prompt_set_distribution}
\centering
\footnotesize
\setlength{\tabcolsep}{4pt}
\renewcommand{\arraystretch}{1.15}
\begin{tabular}{lccl}
\toprule
\textbf{ID} & \textbf{Procedure} & \textbf{Count} & \textbf{Design} \\
\midrule
000--029 & Cholecystectomy      & 30 & 10 actions $\times$ 3 variants \\
030--059 & Gastrectomy          & 30 & 10 actions $\times$ 3 variants \\
060--089 & Hysterectomy         & 30 & 10 actions $\times$ 3 variants \\
090--119 & Intestinal Resection & 30 & 10 actions $\times$ 3 variants \\
120--149 & Nephrectomy          & 30 & 10 actions $\times$ 3 variants \\
150--179 & Prostatectomy        & 30 & 10 actions $\times$ 3 variants \\
\midrule
180--189 & VATS Lobectomy       & 10 & External/generalization mix \\
190--199 & Hepatectomy          & 10 & External/generalization mix \\
\midrule
\textbf{Total} & \textbf{8 procedures} & \textbf{200} & -- \\
\bottomrule
\end{tabular}
\end{table}

The complete list of 200 prompts is released alongside this paper to support reproducibility.

\subsection{Detailed Complete Training Objectives}
\label{app:objectives}

This section provides the complete objectives used to optimize the flow action encoder, text-to-action predictor, and action adapter. The formulations correspond to the selected FLAIR configuration used in the main experiments.

\subsubsection{Flow Action Encoder Objective}

Let a training batch contain $N$ optical-flow representations $\{F_i\}_{i=1}^{N}$. The encoder and decoder produce $\mathbf{z}_i=E_{\phi}(F_i)$ and $\widehat{F}_i=D_{\varphi}(\mathbf{z}_i)$, where each latent token is normalized such that $\|\mathbf{z}_i\|_2=1$. The reconstruction objective is defined as
\begin{equation}
\mathcal{L}_{\mathrm{recon}}
=
\frac{1}{N}
\sum_{i=1}^{N}
\left\|
\widehat{F}_i-F_i
\right\|_2^2.
\label{eq:app_fae_reconstruction}
\end{equation}
The reconstruction term encourages each latent token to retain the information required to recover its corresponding flow representation.

To keep distinct flow observations separable, we apply an instance-discrimination objective. Given the pairwise similarity logits
\begin{equation}
S_{ij}
=
\frac{\mathbf{z}_i^{\top}\mathbf{z}_j}{\tau},
\label{eq:app_instance_logits}
\end{equation}
the instance-discrimination loss is
\begin{equation}
\mathcal{L}_{\mathrm{inst}}
=
-\frac{1}{N}
\sum_{i=1}^{N}
\log
\frac{\exp(S_{ii})}
{\sum_{j=1}^{N}\exp(S_{ij})},
\label{eq:app_instance_loss}
\end{equation}
where $\tau=0.1$ is the similarity temperature. Each flow observation is treated as its own instance, discouraging the encoder from mapping all inputs toward a common latent direction.

We further apply a variance-floor regularizer across the latent dimensions:
\begin{equation}
\mathcal{L}_{\mathrm{var}}
=
\frac{1}{d_z}
\sum_{r=1}^{d_z}
\max
\left(
0,
\gamma-
\operatorname{Std}_{i}
\left[
z_{i,r}
\right]
\right),
\label{eq:app_fae_variance}
\end{equation}
where $z_{i,r}$ is dimension $r$ of token $\mathbf{z}_i$ and $\gamma=0.07$ is the minimum target standard deviation. This term penalizes collapsed dimensions only when their batch-level variation falls below the specified floor and does not encourage unbounded latent variance.

The complete FAE objective is
\begin{equation}
\mathcal{L}_{\mathrm{FAE}}
=
\mathcal{L}_{\mathrm{recon}}
+
0.5\mathcal{L}_{\mathrm{inst}}
+
25.0\mathcal{L}_{\mathrm{var}}.
\label{eq:app_fae_total}
\end{equation}
The selected FAE configuration does not use supervised contrastive learning, domain-adversarial training, covariance penalties, or mean-direction penalties.

\subsubsection{Text-to-Action Prediction Objective}

The FAE and Long-CLIP encoder~\citep{zhang2024long} remain frozen while training the text-to-action predictor. For each prompt--video pair, the predictor produces $\widehat{Z}=[\widehat{\mathbf{z}}_1,\ldots,\widehat{\mathbf{z}}_K]$, while the corresponding real clip provides the frozen FAE target $Z=[\mathbf{z}_1,\ldots,\mathbf{z}_K]$. The predictor objective is
\begin{equation}
\mathcal{L}_{\mathrm{pred}}
=
\frac{1}{K}
\sum_{k=1}^{K}
\left[
1-
\cos
\left(
\widehat{\mathbf{z}}_k,
\mathbf{z}_k
\right)
\right]
+
0.05
\frac{1}{K}
\sum_{k=1}^{K}
\left\|
\widehat{\mathbf{z}}_k-
\mathbf{z}_k
\right\|_2^2.
\label{eq:app_predictor_loss}
\end{equation}
The cosine term aligns the direction of the predicted and flow-derived tokens, while the Euclidean term penalizes element-wise deviations. All $K=80$ positions are predicted simultaneously, and both the predicted and target tokens are $\ell_2$-normalized.

\subsubsection{Action Adapter Objective}

The action adapter is optimized through the forward pass of the frozen, domain-adapted video generator. Given a clean video latent $x_0$, Gaussian noise $\epsilon\sim\mathcal{N}(0,I)$, and interpolation time $u$, the flow-matching path and target velocity are
\begin{equation}
x_u
=
(1-u)x_0+u\epsilon,
\qquad
\mathbf{v}^{\star}
=
\epsilon-x_0.
\label{eq:app_flow_matching_path}
\end{equation}
For an action sequence $Z$, the denoising objective is
\begin{equation}
\mathcal{L}_{\mathrm{den}}(Z)
=
\left\|
D_{\theta}
\left(
x_u,u,\mathbf{c};A_{\omega}(Z)
\right)
-
\mathbf{v}^{\star}
\right\|_2^2,
\label{eq:app_denoising_loss}
\end{equation}
where $D_{\theta}$ is the frozen video denoiser, $\mathbf{c}$ is the backbone-native text condition, and $A_{\omega}(Z)$ denotes the global and temporal conditions produced by the adapter. Although $\theta$ remains frozen, gradients are propagated through the denoiser to update the adapter parameters $\omega$.

To ensure that the adapter uses the supplied action sequence, we compare the matched sequence $Z^{+}$ from the target video with a mismatched sequence $Z^{-}$ from another training clip. Both forward passes share the same noisy latent $x_u$, interpolation time $u$, and text condition $\mathbf{c}$, leaving the action sequence as the only changed condition. The ranking loss is
\begin{equation}
\mathcal{L}_{\mathrm{rank}}
=
\max
\left(
0,
0.02
+
\mathcal{L}_{\mathrm{den}}(Z^{+})
-
\mathcal{L}_{\mathrm{den}}(Z^{-})
\right).
\label{eq:app_ranking_loss}
\end{equation}
This term encourages the matched action sequence to provide a more useful denoising condition than an unrelated motion sequence.

We apply temporal smoothness regularization to the adapter's temporal conditions:
\begin{equation}
\mathcal{L}_{\mathrm{smooth}}
=
\frac{1}{L-1}
\sum_{\ell=2}^{L}
\left\|
\mathbf{b}_{\ell}
-
\mathbf{b}_{\ell-1}
\right\|_2^2.
\label{eq:app_temporal_smoothness}
\end{equation}
This term discourages abrupt changes between neighboring temporal biases.

We also regularize variation in the magnitudes of the temporal conditions:
\begin{equation}
\mathcal{L}_{\mathrm{frame}}
=
\operatorname{Var}_{\ell}
\left(
\left\|
\mathbf{b}_{\ell}
\right\|_2
\right).
\label{eq:app_frame_norm_variance}
\end{equation}
This term prevents individual temporal positions from receiving disproportionately strong injection magnitudes.

The global condition is constrained using an upper-bound penalty:
\begin{equation}
\mathcal{L}_{\mathrm{global}}
=
\max
\left(
0,
\left\|
\mathbf{g}
\right\|_2
-
0.5
\right).
\label{eq:app_global_upper}
\end{equation}
The penalty is active only when the global-condition norm exceeds $0.5$, limiting disruption to the pretrained diffusion-time representation.

The complete condition regularization term is
\begin{equation}
\mathcal{L}_{\mathrm{reg}}
=
0.01\mathcal{L}_{\mathrm{smooth}}
+
0.01\mathcal{L}_{\mathrm{frame}}
+
0.10\mathcal{L}_{\mathrm{global}}.
\label{eq:app_adapter_regularization}
\end{equation}
Finally, the complete action-adapter objective is
\begin{equation}
\mathcal{L}_{\mathrm{adapter}}
=
\mathcal{L}_{\mathrm{den}}(Z^{+})
+
0.20\mathcal{L}_{\mathrm{rank}}
+
\mathcal{L}_{\mathrm{reg}}.
\label{eq:app_adapter_total}
\end{equation}
Adapter optimization uses real flow-derived sequences $Z^{+}$ extracted by the frozen FAE, whereas inference replaces them with the text-predicted prior $\widehat{Z}$.

\subsection{Detailed FLAIR Training Setup and Procedure}
\label{app:training_procedure}


\paragraph{Stage I: Flow-action learning.} The flow action encoder is first trained on real surgical clips to learn a compact motion representation from optical-flow sequences. The decoder is optimized jointly to preserve motion information, while auxiliary objectives encourage the latent space to maintain action-discriminative structure and avoid collapse. After this stage, only the latent action encoder is retained as the motion representation module.

\paragraph{Stage II: Text-to-action prediction.} The flow action encoder is frozen and provides target latent action sequences for paired surgical captions. The text-to-action predictor learns to map surgical prompts into the same latent action space. Since the target latent sequence remains fixed during this stage, the predictor learns the correspondence between textual descriptions and motion priors.

\paragraph{Stage III: Action-adapter optimization.} The action adapter is trained through the frozen video diffusion backbone. The backbone and surgical-domain LoRA remain unchanged, while gradients are propagated through the denoising process to update only the adapter parameters. During training, the adapter receives flow-derived latent actions, whereas inference replaces them with text-predicted latent actions.

\subsection{More Qualitative Results}
\label{sec:supp_more_qualitative}

To further validate the generality of FLAIR's motion improvements across different video diffusion backbones, we provide two additional temporal comparison examples, spanning HunyuanVideo-1.5~\citep{wu2025hunyuanvideo} and CogVideoX-2B~\citep{yang2025cogvideox}, complementing the Wan-based~\citep{wan2025wan} comparison shown in the main paper (Figure~3).

\begin{figure*}[t]
\centering
\includegraphics[width=\textwidth]{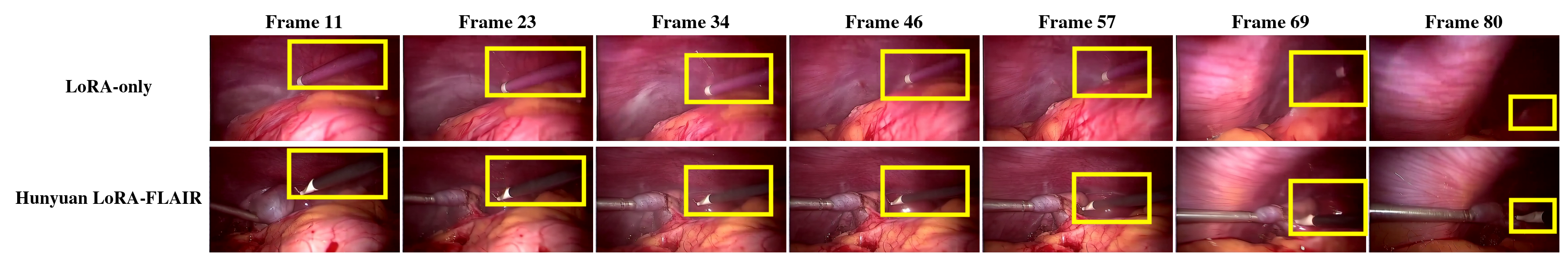}
\caption{Qualitative temporal comparison between HunyuanVideo-1.5 LoRA and HunyuanVideo-1.5 LoRA-FLAIR, with key changes highlighted by bounding boxes. Under identical generation settings, HunyuanVideo-1.5 LoRA-FLAIR produces visually and anatomically more plausible surgical scenes with a consistently visible, stable instrument throughout the sequence, whereas the LoRA-only baseline exhibits drifting instrument motion and localized structural artifacts in later frames. Seven synchronized frames are shown from frame 0 to frame 80.}
\label{fig:supp_hunyuan}
\end{figure*}

\begin{figure*}[t]
\centering
\includegraphics[width=\textwidth]{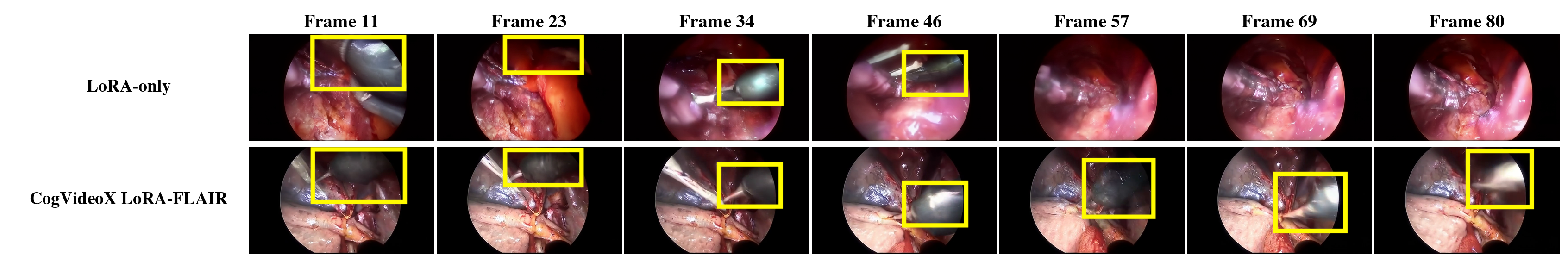}
\caption{Qualitative temporal comparison between CogVideoX-2B LoRA and CogVideoX-2B LoRA-FLAIR, with key changes highlighted by bounding boxes. CogVideoX-2B LoRA-FLAIR maintains a grounded, physically anchored instrument with consistent visibility and a more stable camera view throughout the sequence, while the LoRA-only baseline shows the instrument intermittently disappearing, floating detached from tissue, and increasing viewpoint instability as generation progresses. Seven synchronized frames are shown from frame 0 to frame 80.}
\label{fig:supp_cogvideox}
\end{figure*}

\paragraph{HunyuanVideo-1.5 LoRA-FLAIR}
As shown in Figure~\ref{fig:supp_hunyuan}, HunyuanVideo-1.5 LoRA-FLAIR produces anatomically plausible surgical scenes with a consistently visible, stable instrument across the full 80-frame sequence. In contrast, the LoRA-only baseline exhibits drifting instrument motion and localized structural artifacts in later frames, consistent with its higher artifact-evidence scores on the drift and instrument-deformation axes reported in Main text.

\paragraph{CogVideoX-2B LoRA-FLAIR}
Figure~\ref{fig:supp_cogvideox} shows a similar pattern for CogVideoX-2B. CogVideoX-2B LoRA-FLAIR maintains a grounded, physically anchored instrument with consistent visibility and a more stable camera view throughout the sequence, whereas the LoRA-only baseline shows the instrument intermittently disappearing, floating detached from tissue, and increasing viewpoint instability as generation progresses. These additional examples reinforce the main paper's finding that injecting flow-derived latent action priors improves motion stability and action consistency across multiple base video diffusion architectures, rather than being specific to a single backbone.

\section{Additional Details on SurgMetrics}
\label{app:surgmetrics_details}

This section provides additional details on the degradation design, axis-level supervision, empirical validation, and ablation studies used to construct and validate SurgMetrics. The overall objective is to convert a small set of clean, real laparoscopic clips into a broad and systematically controlled space of surgical-domain failure modes, each targeting one or more clinically meaningful artifact axes, without relying on costly expert annotation of generated video. We first summarize the 21 controlled degradation families and their mapping onto the six prediction heads, then present validation evidence linking SurgMetrics scores to human judgments of generated surgical video quality, and finally report ablation studies examining the contribution of individual design components, including the role of optical-flow-derived motion statistics, to overall evaluator performance.

\subsection{Twenty-one Controlled Degradation Families}
\label{sec:supp_degradations}

To construct axis-specific supervision without relying on expert annotation of generated video, SurgMetrics is trained on 21 controlled synthetic degradation families applied to clean, real laparoscopic clips. Each family's severity is routed, with a family-specific weight, to one or more of the six artifact-specific heads (Semantic, Appearance, Drift, Instrument, Tissue, Tracklet). 

Table~\ref{tab:degradation_definitions} summarizes the meaning and construction of each family, grouped by its primary target axis. Intuitively, the six axes target the recurring, enumerable failure modes of surgical video generation. Semantic Implausibility and Appearance Artifacts capture whether video remains visually and actionally plausible. Drift and Tracklet instability capture whether instruments and tissue remain temporally coherent across frames. Instrument and Tissue Deformation capture whether their local geometry stays physically consistent and assess the cross-frame stability of surgical motion.

Table~\ref{tab:degradation_mapping} reports the full routing-weight matrix from each family to the six heads. As shown in the mapping, most families are routed primarily to a single head, while a subset also contributes secondary supervision to related axes, e.g., the Instrument and Tissue families additionally inform the Tracklet head, reflecting how local instrument or tissue instability naturally manifests as tracklet-level artifacts. When designing the Drift-related families, we deliberately avoided any degradation that penalizes motion purely based on its overall speed, such as treating slower motion as higher quality or faster motion as lower quality. Real surgical videos naturally vary in pacing across surgeons, procedures, and camera viewpoints, so a fixed notion of "correct" speed would not generalize across this variation. Instead, our temporal families (e.g., speed\_distortion, motion\_jerk) specifically target non-uniform, erratic, or discontinuous changes in speed within a clip, rather than penalizing any particular absolute pace.

\begin{figure*}[t]
\centering
\includegraphics[width=0.99\textwidth]{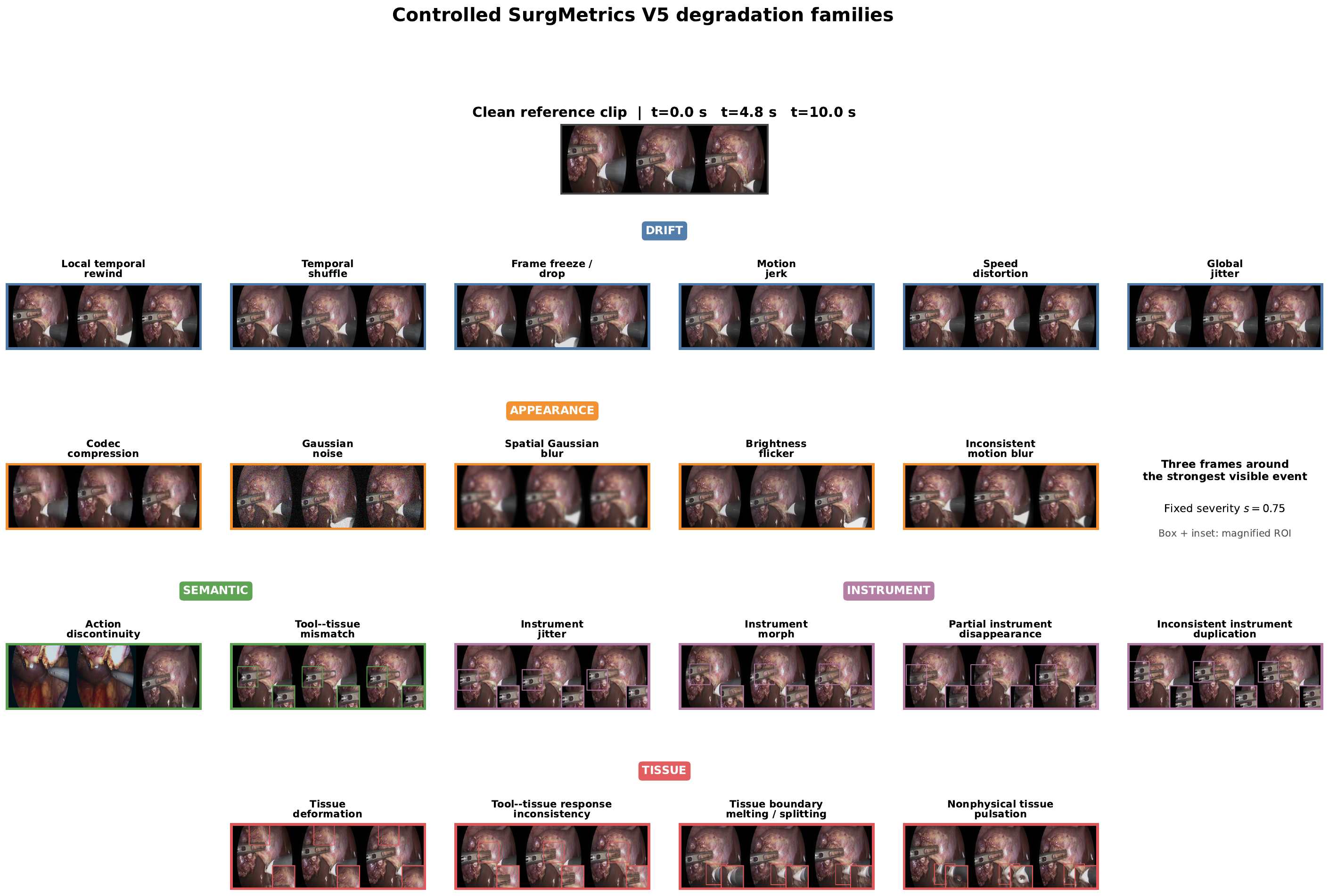}
\caption{Qualitative visualization of the 21 controlled degradation families applied to a representative clean laparoscopic clip, grouped by their primary supervision axis (Drift, Appearance, Semantic, Instrument, and Tissue). Each tile shows three temporally ordered frames rendered at a fixed within-family severity of $s=0.75$. Localized instrument and tissue degradations are shown with magnified regions of interest, while semantic action discontinuity uses a color-matched donor segment. Tracklet is not shown as a separate group because it receives auxiliary supervision from the instrument and tissue families rather than an exclusive degradation family.}
\label{fig:degradation_examples}
\end{figure*}

Figure~\ref{fig:degradation_examples} visualizes the qualitative effect of each of the 21 degradation families applied to a single clean reference frame, illustrating the range of surgical-specific failure modes that SurgMetrics is trained to detect.

\subsection{Validation of SurgMetrics}
\label{sec:surgmetrics_validation}

We validate SurgMetrics along three complementary dimensions: sensitivity to controlled artifacts, generalization beyond the training distribution, and alignment with human perception. SurgMetrics is trained using 100,000 unique source windows sampled from five real surgical datasets, totaling 20,000 real surgical clips, and 21 controlled degradation families (Section~\ref{sec:supp_degradations}). All data are divided at the complete-case level, such that clips from the same surgical procedure cannot appear across training and evaluation splits.

For severity evaluation, we report paired-$\Delta$ Spearman correlation $\rho_{\Delta}$ and clean-versus-degraded AUC. Paired-$\Delta$ correlation measures whether the score increase between a clean clip and its degraded counterpart follows the target degradation severity, thereby reducing variation caused by the original video content.

\begin{table}[t]
\caption{Artifact sensitivity and external generalization of SurgMetrics. Values report mean$\pm$std over three seeds. External evaluation pools SLAM and PhaKIR; single and multi denote one and multiple simultaneous degradation families.}
\label{tab:surgmetrics_validation}
\centering
\footnotesize
\setlength{\tabcolsep}{4pt}
\renewcommand{\arraystretch}{1.15}
\begin{tabular}{@{}lcc@{}}
\toprule
\textbf{Evaluation setting} & \textbf{$\rho_{\Delta}$ $\uparrow$} & \textbf{AUC $\uparrow$} \\
\midrule
Held-out cases (multi) & $0.566{\pm}0.004$ & $0.914{\pm}0.004$ \\
External (single) & $0.731{\pm}0.000$ & $0.924{\pm}0.001$ \\
External (multi) & $0.336{\pm}0.002$ &  $0.743{\pm}0.006$\\
\bottomrule
\end{tabular}

\vspace{2pt}
\end{table}

\paragraph{Artifact sensitivity and external generalization.}
As shown in Table~\ref{tab:surgmetrics_validation}, SurgMetrics reliably distinguishes clean clips from clips containing multiple simultaneous artifacts on previously unseen surgical cases. The high external single-artifact correlation and AUC further show that the learned axes generalize across acquisition conditions and surgical datasets. Performance remains positive for unseen combinations of multiple artifacts, although this setting is more difficult than isolated degradation. The low variation across seeds indicates that these conclusions are not determined by a particular initialization. Complementary results on domain typicality and human-alignment validation are reported in later sections.

\subsubsection{Human-Alignment Validation}
\label{sec:supp_human_alignment}
The evaluation set contains 90 generated laparoscopic videos from nine model configurations, with ten videos per configuration. The same ten prompts are used across all configurations to enable matched comparisons. All videos are anonymized and randomly shuffled, with only the corresponding prompts shown to the evaluators. Four surgeons with specialist qualifications and approximately 7--21 years of clinical experience independently rate each video on a five-point ordinal scale. The evaluation covers visible defects corresponding to the six SurgMetrics artifact dimensions, overall video quality, and the correctness of the instrument--action--target interaction specified by the prompt.

\paragraph{Axis-Level Metric--Human Alignment.}
To evaluate whether SurgMetrics captures native generation failures beyond the synthetic degradations used for training, we compare each SurgMetrics artifact axis directly with the corresponding human rating rather than constructing an aggregate score across heterogeneous axes. Spearman rank correlations range from 0.746 to 0.867 across the six axes, indicating strong positive rank agreement between SurgMetrics and the corresponding human judgments.

\paragraph{Inter-Rater Reliability and Model Dependence.}
We assess the consistency of the human ratings using a two-way random-effects, absolute-agreement ICC for the mean of four raters (ICC(2,4)) \citep{koo2016guideline}. Reliability ranges from 0.721 to 0.832 across the artifact axes, while ICC values for action correctness range from 0.742 to 0.825, indicating generally good inter-rater reliability. We further examine the effect of model identity. The correlation with overall human-rated quality decreases from 0.845 to 0.731 after adjustment, while remaining strong, indicating that the observed association is not solely explained by differences among model configurations.

\paragraph{Action Correctness.}
Action correctness is assessed independently from SurgMetrics by asking the surgeons to evaluate whether the instrument--action--target interaction specified by each prompt is correctly executed. Across all three video backbones, FLAIR receives lower mean interaction-error ratings than the corresponding LoRA baseline, with paired confidence intervals excluding zero. This direct human assessment complements the artifact-oriented SurgMetrics evaluation and supports consistent improvements in action execution across the three backbones.





\subsection{Domain Distance Reference Values}
\label{sec:supp_domain_distance}

To provide practical reference points for interpreting the domain-typicality score, we evaluate the domain-distance branch of SurgMetrics on five datasets spanning a deliberate spectrum from in-domain laparoscopic video to fully out-of-domain general video, using 100 clips per dataset. Table~\ref{tab:domain_distance_reference} reports the mean robust $z$-score and the AUROC of separating each dataset from in-domain unseen test clips.

\begin{table}[t]
\caption{Domain-distance reference values across datasets spanning in-domain, semi-in-domain, and out-of-domain video, using 100 clips per dataset. Higher values indicate greater deviation from the training laparoscopic distribution.}
\label{tab:domain_distance_reference}
\centering
\footnotesize
\setlength{\tabcolsep}{6pt}
\renewcommand{\arraystretch}{1.15}
\begin{tabular}{lcc}
\toprule
\textbf{Dataset} & \textbf{Mean robust $z$} $\downarrow$ & \textbf{AUROC vs.\ in-domain} $\uparrow$ \\
\midrule
In-domain unseen test & 0.5614 & 0.5000 \\
PhaKIR & 1.8538 & 0.7646 \\
SLAM & 1.9552 & 0.7933 \\
OphNet & 3.6019 & 0.9547 \\
SurgVU24 & 4.9917 & 0.9923 \\
UCF101 & 5.8199 & 0.9957 \\
\bottomrule
\end{tabular}
\end{table}

These results order datasets exactly as expected given their surgical relatedness. PhaKIR and SLAM, both near-domain laparoscopic datasets, remain closest to the training distribution (robust $z$ of 1.85 and 1.96, still separable from in-domain unseen test but far below the datasets that follow). OphNet, a semi-related surgical domain (ophthalmic surgery, sharing instrument-tissue interaction structure but differing anatomy and imaging setup), sits at an intermediate distance (3.60). SurgVU24, a robotic-surgery dataset with substantially different visual and platform characteristics, and UCF101, a fully unrelated general-domain action-recognition dataset, are pushed furthest from the training distribution (4.99 and 5.82, respectively), confirming that the domain-distance branch correctly separates genuinely out-of-surgical-domain content from surgical video that merely differs in sub-specialty.

Based on this progression and consistent with the domain-distance threshold of 4 used in the main paper to exclude out-of-domain clips from SurgMetrics comparisons, we suggest the following coarse interpretation bands: values below approximately 2 indicate in-domain or near-domain laparoscopic content (e.g., PhaKIR, SLAM); values in the intermediate range (roughly 2--4) indicate semi-related surgical domains that are typologically surgical but visually or anatomically distinct (e.g., OphNet); and values above 4 indicate content that has departed the surgical laparoscopic distribution to a degree where SurgMetrics artifact scores are no longer meaningful for comparison (e.g., SurgVU24, UCF101), consistent with our exclusion criterion in the main paper.

It is important to emphasize that a higher robust $z$-score reflects greater distributional distance from the training domain, not more severe generation artifacts; domain typicality and the six artifact-specific heads are deliberately reported as independent, non-substitutable signals. A clip can be domain-typical while still exhibiting severe artifacts, or domain-atypical while visually clean, and conflating the two would misrepresent what each score is designed to measure.

\subsection{Ablation Study on SurgMetrics}
\label{sec:supp_ablation_surgmetrics}

SurgMetrics fuses two input branches before predicting the six artifact heads: a SurgViSTA~\citep{yang2026large} visual-feature leg and a 25-dimensional Farneback optical-flow leg~\citep{farneback2003two}. To isolate each branch's contribution, we compare Full (both legs fused), SurgViSTA-only, and Optical-only, all trained under identical data splits, epochs, seeds (42/43/44), and evaluation cohorts (1,500 held-out multi-degradation records, 1,122 external records, 1,200 clean validation records for threshold calibration).

\begin{table}[t]
\centering
\scriptsize
\setlength{\tabcolsep}{8pt}
\renewcommand{\arraystretch}{1.15}
\begin{tabular}{lccccccc}
\toprule
& & & & \multicolumn{2}{c}{\textbf{Held-out}} & \multicolumn{2}{c}{\textbf{External}} \\
\cmidrule(lr){5-6} \cmidrule(lr){7-8}
\textbf{Configuration} & \textbf{Params} & \textbf{Test Acc $\uparrow$} & \textbf{Internal $\rho$ $\uparrow$} & \textbf{multi $\rho$ $\uparrow$} & \textbf{AUC $\uparrow$} & \textbf{single $\rho$ $\uparrow$} & \textbf{single AUC $\uparrow$} \\
\midrule
Full & 354{,}502 & \textbf{0.8250} & \textbf{0.5436} & \textbf{0.5661} & \textbf{0.9144} & \textbf{0.7307} & \textbf{0.9243} \\
SurgViSTA-only & 345{,}478 & 0.7917 & 0.5144 & 0.5278 & 0.9029 & 0.5934 & 0.8556 \\
Optical-only & 92{,}102 & 0.6569 & 0.2779 & 0.2452 & 0.6376 & 0.5752 & 0.7729 \\
\bottomrule
\end{tabular}
\caption{Ablation over SurgMetrics input branches across adapter test, internal, held-out, and external evaluation cohorts. All values are averaged over seeds 42/43/44.}
\label{tab:ablation_overall}
\end{table}

As shown in Table ~\ref{tab:ablation_overall}, full achieves the best score on every aggregate metric. Relative to SurgViSTA-only, adding the optical-flow leg costs only 9{,}024 extra parameters yet yields consistent gains across all aggregate metrics, most notably +0.1373 Spearman on the external single-family cohort, while relative to Optical-only, adding the SurgViSTA leg yields substantially larger gains (e.g., +0.3209 held-out multi Spearman, +0.2768 held-out AUC), confirming SurgViSTA as the primary semantic/visual backbone that the optical leg alone cannot replace.

\begin{table}[t]
\caption{Drift-axis performance under each ablation configuration, the axis most affected by the optical-flow leg.}
\label{tab:ablation_drift}
\centering
\footnotesize
\setlength{\tabcolsep}{6pt}
\renewcommand{\arraystretch}{1.15}
\begin{tabular}{lcc}
\toprule
\textbf{Configuration} & \textbf{Adapter test} & \textbf{Internal} \\
& \textbf{Drift Acc} & \textbf{Drift $\rho$} \\
\midrule
Full & \textbf{0.9500} & \textbf{0.6106} \\
SurgViSTA-only & 0.7250 & 0.4375 \\
Optical-only & 0.5333 & 0.5235 \\
\bottomrule
\end{tabular}
\end{table}

Table ~\ref{tab:ablation_drift} isolates the clearest per-axis signal, Drift. SurgViSTA-only trails Full by 0.2250 in adapter test accuracy and 0.1731 in internal Spearman, while Optical-only alone retains a comparatively strong Drift correlation (0.5235) despite far fewer parameters and no visual semantics. This indicates that optical features primarily complement SurgViSTA by supplying temporal-motion evidence, with the strongest gain concentrated in drift-related evaluation, whereas SurgViSTA alone cannot fully capture motion-timing artifacts and optical features alone cannot sustain full six-axis evaluation capability.

\section{Detailed Training Setup}
\label{app:method_details}

\subsection{Detailed LoRA training Setup}
\label{sec:detailed_lora_training_setup}

We adapted three pretrained text-to-video models---Wan2.1-T2V-1.3B~\citep{wan2025wan}, HunyuanVideo-1.5 T2V (8.3B)~\citep{wu2025hunyuanvideo}, and CogVideoX-2B~\citep{yang2025cogvideox}---using low-rank adaptation (LoRA). For a frozen linear projection $W$, LoRA parameterizes the update as $W' = W + (\alpha/r)BA$, where $r$ is the rank. Only the LoRA parameters were optimized, while the pretrained video backbone, VAE, and text encoders remained frozen. We used a LoRA rank of $r=32$ and scaling parameter $\alpha=1$ for all three models. The training videos were paired with detailed English captions describing the surgical instruments, actions, anatomy, and visual context.

\paragraph{Wan2.1.}
We fine-tuned the Wan2.1-T2V-1.3B model using the Wan LoRA architecture provided by Musubi Tuner. The LoRA parameters were randomly initialized and trained directly from scratch on top of the pretrained Wan backbone; no previously trained adapter or optimizer state was used. LoRA layers were applied to the linear projections in the transformer's attention blocks. Training was performed for four epochs using six GPUs with data parallelism, a per-GPU batch size of 1, and gradient accumulation over two steps, resulting in an effective batch size of 12. We used 8-bit AdamW with a constant learning rate of $2\times10^{-4}$, a maximum gradient norm of 1.0, and random seed 42. The majority of training samples contained 81 frames at $480\times832$ pixels, while short clips contained 33 frames. Shifted timestep sampling with a discrete flow shift of 3.0 was used. Training was conducted with BF16 mixed precision, FP8 backbone weights, scaled-dot-product attention, and gradient checkpointing.

\paragraph{HunyuanVideo-1.5.}
We fine-tuned the 720p T2V variant of HunyuanVideo 1.5 using its dedicated LoRA architecture in Musubi Tuner. This model employs a multimodal diffusion transformer together with a video VAE and Qwen2.5-VL and ByT5 text-conditioning branches. The LoRA parameters were initialized from scratch and inserted into the model's multimodal transformer blocks. Training was performed for three epochs using six-GPU data parallelism, a per-GPU batch size of 1, and two-step gradient accumulation, giving an effective batch size of 12. We used 8-bit AdamW with a constant learning rate of $1\times10^{-4}$, gradient clipping at 1.0, and seed 42. Videos were trained primarily as 81-frame clips at $320\times576$ pixels, with 33-frame windows used for shorter clips. We adopted shifted timestep sampling without additional loss weighting and used a discrete flow shift of 2.0. BF16 mixed precision, scaled FP8 backbone weights, scaled-dot-product attention, and gradient checkpointing were enabled during training.

\paragraph{CogVideoX.}
We fine-tuned CogVideoX-2B using the Hugging Face FineTrainers architecture, which integrates the Diffusers implementation of CogVideoX with PEFT-based LoRA training. CogVideoX uses a latent video transformer conditioned on T5 text features and a 3D VAE for video encoding and decoding. LoRA was applied to the query, key, value, and output projections of the transformer attention layers. Training used six-GPU data parallelism with a per-GPU batch size of 1 and two-step gradient accumulation, yielding an effective batch size of 12. The videos were resampled to 16 fps and organized into 81-frame and 33-frame resolution buckets. We used AdamW with a constant learning rate of $2\times10^{-4}$, gradient clipping at 1.0, and seed 42. The model was trained in BF16 with gradient checkpointing, and the 6,500-step LoRA checkpoint was selected for evaluation.

\subsection{Detailed FAE Training Setup}
\label{sec:supp_fae_setup}

The final Flow Action Encoder (FAE) uses checkpoint step 10,000, with a 128-dimensional latent and 128$\times$128 training resolution. Input flow images are stored as 3$\times$256$\times$256 uint8, normalized to $[-0.5, 0.5]$, and bilinearly resized to 3$\times$128$\times$128. Each 80-position flow window is sampled as a pair of positions 40 apart, $i \sim \mathrm{Uniform}\{0,\dots,39\}$ and $i+40$, per training step.

\paragraph{Architecture.}
The encoder is a convolutional stack (3$\times$128$\times$128 $\to$ 24$\to$24$\to$48$\to$96$\to$192$\to$384$\to$768 channels, halving spatial resolution at each stage down to 1$\times$1) followed by a linear projection to 128 dimensions and L2 normalization; the decoder mirrors this structure in reverse, ending with a $3\times3$ convolution and a $\tanh/2$ output. The full model has 7,399,103 parameters, and encoding a full window yields $Z \in \mathbb{R}^{80\times128}$.

\paragraph{Optimization.}
We train for 10,000 steps on 4 GPUs (DDP) with AdamW (lr 1e-4, weight decay 0.01, BF16 precision, gradient clipping 20), 32 windows per GPU (128 global windows / 256 flow images per step), seed 42, validating every 1,000 steps. Training uses hierarchical sampling (dataset $\to$ action $\to$ case $\to$ window) across five surgical datasets, with dataset sampling probabilities of 0.306 (Cholec80~\citep{twinanda2016endonet}), 0.252 (M2CAI16), 0.206 (AutoLaparo~\citep{wang2022autolaparo}), 0.211 (HeiCo), and 0.024 (SurgicalActions160~\citep{SurgicalActions160}), over 43,013 training / 8,256 validation / 7,266 held-out test windows. Checkpoint step 10,000 is selected based on downstream motion utility measured by a frozen world-model probe, rather than reconstruction error alone.

\subsection{Detailed Predictor Training Setup}
\label{sec:supp_predictor_setup}

The text-to-action predictor pairs a frozen Long-CLIP-L text encoder with a non-autoregressive Transformer decoder, trained to regress the frozen FAE latent sequence. Input captions are tokenized (max 248 tokens) and pooled by frozen Long-CLIP-L into a 768-dim embedding; the target is the frozen FAE $z$ sequence $[80,128]$. Only the predictor is trained; Long-CLIP remains entirely frozen.

\paragraph{Architecture.}
The pooled text embedding is projected (768$\to$256) into a single context token, which conditions 80 learned temporal queries through a 4-layer Transformer decoder (hidden dim 256, 4 heads, FFN dim 1024, dropout 0.1); a final linear layer (256$\to$128) with per-position L2 normalization produces the predicted sequence $\hat{Z} \in \mathbb{R}^{80\times128}$. All 80 positions are predicted in parallel, not autoregressively.

\paragraph{Optimization.}
We train on a single GPU with AdamW (fixed lr 3e-4, weight decay 1e-4), batch size 64 (validation batch 128), for up to 10,000 steps (gradient clipping 1.0, seed 42), validating and saving every 500 steps using the same train/validation/test window split as the FAE (43,013 / 8,256 / 7,266). Model selection uses validation frame-level cosine similarity; the best checkpoint occurs at step 500.

\subsection{Detailed Adapter Training Setup}
\label{sec:supp_adapter_setup}

All three backbones (Wan2.1-T2V-1.3B, HunyuanVideo 1.5, CogVideoX-2B) share a common motion-side interface: the FAE/predictor latent $[B,80,128]$ is mean-pooled every 4 positions to $[B,20,128]$, followed by LayerNorm and a SiLU MLP (128$\to$512$\to$512) producing a global condition and a temporal frame bias. Adapters are trained using real $z$ extracted from real optical flow, but at inference use predicted $z$ generated from the caption; the video backbone and surgical LoRA are frozen throughout, with only the adapter updated. All three adapters inject conditioning at the input level (global/time-bias addition), rather than through per-block cross-attention.

\paragraph{Shared optimization settings.}
All adapters are trained with AdamW (lr 1e-4, weight decay 1e-4), per-GPU batch size 1, gradient clipping 1.0, BF16 precision, and seed 42, using 1,024 training samples (128 validation samples for Hunyuan/CogVideoX).

\paragraph{Backbone-specific configurations.}
Table~\ref{tab:adapter_configs} summarizes the injection dimensionality, trainable parameters, training budget, and deployment settings for each backbone. The Wan adapter (1.9M parameters) injects a 1536-dim global condition and $[B,20,1536]$ temporal bias into the time embedding and patch-embedding features, trained for 1,000 steps on a single GPU at 480$\times$832 resolution; step 500 is selected over step 1,000 for deployment (scale 0.30). The Hunyuan adapter (2.4M parameters) injects a 2048-dim condition into \texttt{transformer.time\_in} and \texttt{transformer.img\_in}, with the 20-position temporal bias linearly interpolated to the 21 VAE latent time positions; it is trained on 4 GPUs (DDP) at 320$\times$576 resolution for the full 1,000 steps (deployment scale 0.50). The CogVideoX adapter (1.6M parameters) injects a 512-dim global condition and $[B,20,1920]$ temporal bias into the video-token portion of the patch embedding only (excluding text tokens); the final paper-facing branch pairs the adapter with a surgical LoRA (rank 32, effective scale 1/32), trained on 8 GPUs (DDP) at 480$\times$832 resolution, with the final deployed checkpoint at step 400 (scale 0.50).

\begin{table*}[t]
\caption{Summary of adapter configurations and deployment settings across the three video backbones.}
\label{tab:adapter_configs}
\centering
\footnotesize
\setlength{\tabcolsep}{4pt}
\renewcommand{\arraystretch}{1.15}
\begin{tabular}{lccccc}
\toprule
\textbf{Backbone} & \textbf{Params} & \textbf{Training budget} & \textbf{Resolution} & \textbf{Selected step} & \textbf{Adapter scale} \\
\midrule
Wan2.1-T2V-1.3B & 1,904,896 & 1k / 1 GPU & 480$\times$832 & 500 & 0.30 \\
HunyuanVideo 1.5 & 2,430,208 & 1k / 4 GPU (DDP) & 320$\times$576 & 1,000 & 0.50 \\
CogVideoX-2B & 1,576,576 & 1k / 8 GPU (DDP) & 480$\times$832 & 400 & 0.50 (LoRA 1/32) \\
\bottomrule
\end{tabular}
\end{table*}

\subsection{Detailed SurgMetrics Training Setup}
\label{sec:supp_training_setup}

SurgMetrics is trained in two stages: a SurgViSTA backbone adaptation stage, followed by a lightweight fusion model trained on frozen, cached features. The system is not trained end-to-end; SurgViSTA is first adapted and frozen before the fusion MLP is trained on top of its cached outputs.

\paragraph{Stage 1: SurgViSTA backbone fine-tune.}
Starting from the pretrained SurgViSTA backbone, we fine-tune only the last 4 Transformer blocks and \texttt{fc\_norm}, keeping earlier blocks frozen, with a temporary 768$\to$256$\to$6 prediction head (GELU, dropout 0.2) used solely for adaptation. Training uses AdamW with a backbone learning rate of 1e-5 and head learning rate of 3e-4 (no scheduler), weight decay 1e-4, BF16 autocast, gradient clipping 1.0, and Smooth L1 loss ($\beta{=}0.1$), for 6 epochs at seed 42. Training follows a curriculum: the first 2 epochs use only clean and single-degradation records (20,400 total), after which all 30,000 training records (clean, single, double, triple degradations) are used from epoch 3 onward. Effective batch size varies with available GPUs (5 for epoch 1, 8 for epochs 2--6, via PyTorch DDP/NCCL), with total training time of approximately 11.5 hours. Model selection uses mean-axis Spearman correlation across the six heads on a fixed 6,000-record validation set; epoch 5 (0.5066) is selected as the best checkpoint, narrowly outperforming epoch 3 (0.4983) and epoch 6 (0.4999).

\paragraph{Stage 2: Fusion model training.}
The fusion model takes frozen 768-dim SurgViSTA features and 25-dim Farneback optical-flow features as input, projecting each through a small branch (SurgViSTA: 768$\to$256; Optical: 25$\to$32, both with GELU) before concatenation (288-dim) and a shared MLP (288$\to$256$\to$128, GELU, dropout 0.2), followed by six independent heads (128$\to$64$\to$1, GELU, dropout 0.1) producing raw, unbounded scores for Semantic, Appearance, Drift, Instrument, Tissue, and Tracklet. With only 354,502 trainable parameters, training is fast (10--14 seconds/epoch) using cached features. We train for 30 epochs with AdamW, initial learning rate 3e-4 with cosine annealing to 0, weight decay 1e-4, batch size 256, and gradient clipping 5.0; ranking loss is disabled (weight 0). A curriculum similarly restricts the first 5 epochs to clean and single-degradation records (68,000) before introducing the full 100,000-record set (including double and triple degradations) from epoch 6 onward. The primary model uses seed 42, with seeds 43 and 44 used for reproducibility checks; early stopping (patience 6) is used during development but disabled for the final release run to guarantee all 30 epochs complete.

\paragraph{Axis-balanced loss.}
Since positive prevalence varies substantially across heads (from 16.5\% for Semantic to 37.4\% for Tracklet), we apply inverse-positive-frequency weighting, $w_a \propto 1/\max(\text{prevalence}_a, 0.05)$, normalized to a mean of 1 and clipped to $[0.5, 3.0]$, computed using training-set statistics only. The overall objective is
\[
\mathcal{L} = \frac{1}{NA}\sum_{n=1}^{N}\sum_{a=1}^{A} w_a \, \mathrm{SmoothL1}(p_{n,a}, y_{n,a}; \beta{=}0.1).
\]

\paragraph{Data and severity.}
All splits are performed at the case level (split seed 43) across five source datasets (Cholec80 action, AutoLaparo, M2CAI16, SurgicalActions160, HeiCo), ensuring no procedure appears across train/validation/test. From 20,000 unique real source windows, each augmented with 5 degradation variants, we obtain 100,000 training records. Training severity is sampled continuously, $s \sim \mathcal{U}(0.15, 1.00)$, while validation uses fixed severity levels (0.00, 0.25, 0.50, 0.75, 1.00) over 1,200 unique source windows (6,000 records total). When multiple degradations co-occur, per-axis labels are aggregated via noisy-OR, $y_a = 1 - \prod_i (1 - s_i w_{i,a})$, consistent with the routing weights in Table~\ref{tab:degradation_definitions}.

\subsection{Detailed Inference Setup}
\label{sec:supp_inference_setup}

All quantitative results are generated using a shared prompt set of 200 surgical prompts (\texttt{prompt\_set\_v2}), with every video generated at 81 frames and seed 42. Across all three backbones, the three FLAIR variants (Base-only, LoRA-only, LoRA-FLAIR) share identical resolution, frame count, FPS, denoising steps, guidance scale, seed, and LoRA weights within each backbone; the only difference introduced by LoRA-FLAIR relative to LoRA-only is the addition of the action adapter and the predicted motion latent. All three backbones further condition on the same cached $[80,128]$ prompt-$z$ produced by the shared Long-CLIP predictor. Backbone-native settings such as resolution, FPS, denoising steps, guidance scale, scheduler, LoRA scale, and adapter scale otherwise follow each backbone's own validated configuration, summarized in Table~\ref{tab:inference_setup}.

\begin{table*}[t]
\caption{Inference configuration for the 200-prompt evaluation set across the three video backbones. Prompt set, frame count, and seed are shared; all other settings follow each backbone's validated native configuration.}
\label{tab:inference_setup}
\centering
\footnotesize
\renewcommand{\arraystretch}{1.15}
\begin{tabular}{lccc}
\toprule
\textbf{Parameter} & \textbf{Wan2.1-1.3B} & \textbf{HunyuanVideo 1.5} & \textbf{CogVideoX-2B} \\
\midrule
Resolution (W$\times$H) & 832$\times$480 & 576$\times$320 & 832$\times$480 \\
Frames & 81 & 81 & 81 \\
FPS & 16 & 24 & 16 \\
Denoising steps & 20 & 30 & 50 \\
Guidance scale & 5.0 & 6.0 & 6.0 \\
Flow shift & 5.0 & 2.0 & -- \\
Scheduler & Flow UniPC & Native flow-matching sampler & DDIM \\
Seed & 42 & 42 & 42 \\
LoRA scale & 1.0 & 1.0 & 1/32 \\
Adapter checkpoint & Step 500 & Step 1,000 & Step 400 \\
Adapter scale & 0.30 & 0.50 & 0.50 \\
Backbone precision & FP8 & FP8-scaled & BF16 \\
Prompt-$z$ source & \multicolumn{3}{c}{Shared $[80,128]$ cache from Long-CLIP predictor} \\
\bottomrule
\end{tabular}
\end{table*}

\paragraph{Wan2.1-1.3B.}
We use Wan2.1-T2V-1.3B with the FlowUniPCMultistepScheduler, 20 denoising steps, guidance scale 5.0, and flow shift 5.0, at 832$\times$480 resolution with torch attention and FP8 backbone precision; the surgical LoRA is applied at multiplier 1.0, and the trained adapter (step 500) is applied at scale 0.30. No explicit negative prompt is passed, so the backbone's default negative prompt is used.

\paragraph{HunyuanVideo-1.5.}
We use HunyuanVideo-1.5 720p T2V at 576$\times$320 resolution, 24 FPS, 30 denoising steps, guidance scale 6.0, and flow shift 2.0, with torch attention, FP8-scaled backbone precision, and inference block swap disabled (block swapping is used only during training for memory management, not at inference). The surgical LoRA is applied at multiplier 1.0, the trained adapter (step 1,000) at scale 0.50, and the negative prompt is left empty.

\paragraph{CogVideoX-2B.}
We use CogVideoX-2B with the CogVideoXDDIMScheduler, 50 denoising steps, guidance scale 6.0, and BF16 pipeline precision, at 832$\times$480 resolution and 16 FPS, with dynamic CFG disabled and no negative prompt. The epoch-3 surgical LoRA is applied at scale 1/32, and the trained adapter (step 400) is applied at scale 0.50.

\subsection{Detailed implementation of general metrics}
\label{sec:supp_general_metrics}
We evaluate the generated videos using five general metrics: VideoCLIP score, Fr\'echet Inception Distance (FID)~\citep{heusel2017gans}, Fr\'echet Video Distance (FVD)~\citep{unterthiner2019fvd}, temporal intersection over union (TI), and area fluctuation (AF). The same frozen preprocessing, model checkpoints, and reference distribution are used for all generation methods.

\paragraph{VideoCLIP score.}
VideoCLIP evaluates the semantic alignment between a generated video and its conditioning prompt. Each video is decoded at 30 frames per second, resized while preserving its aspect ratio, and center-cropped to $224\times224$. The official VideoCLIP video and text encoders are then used to obtain pooled
representations $\mathbf{v}_i$ and $\mathbf{t}_i$. The score for video $i$ is the native VideoCLIP retrieval score
\begin{equation}
    s_i^{\mathrm{VC}}=\mathbf{v}_i^{\top}\mathbf{t}_i.
\end{equation}
We report the mean score over all videos. A higher VideoCLIP score indicates better semantic consistency between the generated content and the input prompt.

\paragraph{FID and FVD.}
FID and FVD compare the generated-video distribution with a fixed real
surgical-video distribution. We construct a unified reference containing 400
held-out real surgical clips, each with 81 frames, and use the same reference
statistics for every generator. Both distances follow the Fr\'echet distance
between the real and generated feature distributions:
\begin{equation}
    d_{\mathrm{F}}
    = \lVert\boldsymbol{\mu}_{r}-\boldsymbol{\mu}_{g}\rVert_2^2
    + \operatorname{Tr}\!\left(
      \boldsymbol{\Sigma}_{r}+\boldsymbol{\Sigma}_{g}
      -2(\boldsymbol{\Sigma}_{r}\boldsymbol{\Sigma}_{g})^{1/2}
      \right),
\end{equation}
where $\boldsymbol{\mu}$ and $\boldsymbol{\Sigma}$ denote the feature mean and
covariance.

FID is computed using \texttt{pytorch-fid==0.3.0} and 2048-dimensional
Inception-v3 pool3 features. We uniformly sample six frames at indices
$\{0,16,32,48,64,80\}$ from each video. FID primarily measures frame-level
visual fidelity and diversity, such as the realism of anatomy, instruments,
texture, and color. FVD is computed using \texttt{cd-fvd==0.1.1} and a
Kinetics-400 pretrained I3D encoder~\citep{carreira2017quo}, with all 81 frames used jointly. It
therefore measures both visual realism and temporal dynamics. Lower FID and
FVD values indicate that the generated distribution is closer to the real
surgical-video distribution.

\paragraph{Temporal intersection over union.}
TI measures the frame-to-frame spatial consistency of surgical instruments.
Given the unioned binary instrument mask $M_t$ at frame $t$, TI is defined as
\begin{equation}
    \operatorname{TI}(V)
    = \frac{1}{|\mathcal{P}|}\sum_{t\in\mathcal{P}}
      \frac{|M_{t-1}\cap M_t|}{|M_{t-1}\cup M_t|},
\end{equation}
where $\mathcal{P}$ contains adjacent frame pairs for which at least one mask
is non-empty. Pairs with two empty masks are excluded. A higher TI indicates
better temporal continuity, while a lower value suggests abrupt displacement,
deformation, disappearance, or temporal flickering of instruments.

\paragraph{Area fluctuation.}
AF measures abrupt changes in the visible instrument area. Let
$A_t=\sum_x M_t(x)$ denote the number of foreground pixels at frame $t$. AF is
computed as
\begin{equation}
    \operatorname{AF}(V)
    = \frac{1}{T-1}\sum_{t=1}^{T-1}
      \frac{|A_t-A_{t-1}|}{\max(A_{t-1},1)}.
\end{equation}
A lower AF indicates smoother and more stable temporal evolution of instrument size, whereas a higher AF reflects sudden scale changes, shape instability, or instrument appearance and disappearance. Notably, in the comparison of FLAIR models with the LoRA-only models and the base models, CogVideoX-2B LoRA showed a significantly abnormal AF value even when a certain surgical environment and instruments were already apparent. This illustrates the inadequacy of general metrics in the surgical field and further emphasizes the importance of Surgmetrics.

\begin{table*}[t]
\caption{Definitions of the 21 controlled degradation families used to construct SurgMetrics training supervision, grouped by primary target axis.}\label{tab:degradation_definitions}
\centering
\scriptsize
\renewcommand{\arraystretch}{0.99}
\begin{tabular}{p{0.24\textwidth}p{0.63\textwidth}p{0.09\textwidth}}
\toprule
\textbf{Degradation} & \textbf{Description} & \textbf{Head} \\
\midrule
\multicolumn{3}{l}{\textit{A. Drift: temporal and global-motion degradations}} \\
\midrule
\degname{local temporal rewind} & Reverses or loops a local temporal segment (8\%--38\% of the clip), producing action rewind or repetition. & Drift \\
\degname{temporal shuffle} & Swaps adjacent short temporal blocks while preserving intra-block motion order, disrupting local event ordering. & Drift \\
\degname{frame freeze drop} & Replaces random frames with the preceding frame, producing freezing, duplication, and discontinuous jumps after drops. & Drift \\
\degname{motion jerk} & Applies stride-and-hold sampling within a short segment, producing sudden acceleration, pauses, and jerks. & Drift \\
\degname{speed distortion} & Applies non-linear temporal resampling across the clip, causing gradual speed-up or slow-down instead of uniform playback. & Drift \\
\degname{global jitter} & Applies frame-wise random translation and rotation to the entire frame, simulating non-physical global camera shake (distinct from local instrument jitter). & Drift \\
\midrule
\multicolumn{3}{l}{\textit{B. Appearance: visual rendering degradations}} \\
\midrule
\degname{codec compression} & Re-encodes the clip with H.264 at increasing CRF and shortened GOP, producing blocking artifacts and detail loss. & Appearance \\
\degname{gaussian noise} & Adds pixel-wise Gaussian noise to RGB frames (standard deviation approximately 4--40). & Appearance \\
\degname{spatial gaussian blur} & Applies isotropic Gaussian blur per frame, simulating defocus or spatial detail loss (distinct from directional motion blur). & Appearance \\
\degname{brightness flicker} & Perturbs per-frame brightness using smooth random and periodic terms, producing unstable exposure/illumination across frames. & Appearance \\
\degname{excessive or inconsistent motion blur} & Applies directional motion blur to a subset of frames with varying direction and intensity, simulating excessive or temporally inconsistent blur. & Appearance \\
\midrule
\multicolumn{3}{l}{\textit{C. Semantic: action and interaction degradations}} \\
\midrule
\degname{semantic action discontinuity} & Replaces a local segment (8\%--43\% of the clip) with a color-matched donor segment from another surgical clip, producing abrupt action/scene change. & Semantic \\
\degname{semantic tool tissue interaction mismatch} & Uses an instrument mask to relocate the instrument away from its original contact point while inpainting the vacated region, producing an implausible instrument-tissue relationship. & Semantic \\
\midrule
\multicolumn{3}{l}{\textit{D. Instrument: instrument-level degradations}} \\
\midrule
\degname{instrument jitter} & Moves only the masked instrument region with random local jitter while inpainting the original location, keeping the background stable. & Instrument \\
\degname{instrument morph} & Applies time-varying scaling, shearing, and non-rigid affine transforms centered on the instrument, periodically distorting its length, width, or contour. & Instrument \\
\degname{instrument partial disappearance} & Inpaints part or all of the instrument over a continuous interval with fade in/out intensity, causing local disappearance and reappearance. & Instrument \\
\degname{temporally inconsistent instrument duplication} & Creates an offset, wobbling duplicate of the instrument that appears, moves, and vanishes briefly. & Instrument \\
\midrule
\multicolumn{3}{l}{\textit{E. Tissue: tissue-level degradations}} \\
\midrule
\degname{tissue deformation} & Generates a temporally correlated local deformation field on tissue regions away from the instrument, producing sustained non-physical warping while preserving the instrument region. & Tissue \\
\degname{physical tool tissue response inconsistency} & Estimates the contact region from instrument-tip motion but drives tissue response in an incorrect or opposite direction, simulating physically inconsistent tool-tissue reaction. & Tissue \\
\degname{tissue boundary melting or splitting} & Separates or compresses a salient non-instrument tissue boundary along its normal direction and reconnects it, producing melting, splitting, or discontinuous boundary evolution. & Tissue \\
\degname{nonphysical tissue pulsation} & Applies radial expansion/contraction on tissue regions away from the instrument without contact or physiological basis, producing periodic pulsation. & Tissue \\
\bottomrule
\end{tabular}

\end{table*}

\begin{table*}[t]
\caption{Routing-weight matrix from the 21 degradation families to the six SurgMetrics artifact heads. Weights denote label-routing strength used in the noisy-OR aggregation (Eq.~12 in the main paper).}\label{tab:degradation_mapping}
\centering
\footnotesize
\setlength{\tabcolsep}{5pt}
\renewcommand{\arraystretch}{1.05}
\begin{tabular}{p{0.30\textwidth}cccccc}
\toprule
\textbf{Family} & \textbf{Sem.} & \textbf{App.} & \textbf{Drift} & \textbf{Instr.} & \textbf{Tissue} & \textbf{Track.} \\
\midrule
\degname{local temporal rewind} & 0.35 & -- & 1.00 & -- & -- & -- \\
\degname{temporal shuffle} & -- & -- & 0.75 & -- & -- & -- \\
\degname{frame freeze drop} & -- & -- & 0.80 & -- & -- & -- \\
\degname{motion jerk} & -- & -- & 0.80 & -- & -- & -- \\
\degname{speed distortion} & -- & -- & 0.80 & -- & -- & -- \\
\degname{global jitter} & -- & -- & 0.90 & -- & -- & -- \\
\degname{codec compression} & -- & 1.00 & -- & -- & -- & -- \\
\degname{gaussian noise} & -- & 1.00 & -- & -- & -- & -- \\
\degname{spatial gaussian blur} & -- & 1.00 & -- & -- & -- & -- \\
\degname{brightness flicker} & -- & 1.00 & -- & -- & -- & -- \\
\degname{excessive or inconsistent motion blur} & -- & 1.00 & -- & -- & -- & -- \\
\degname{semantic action discontinuity} & 1.00 & -- & 0.35 & -- & -- & -- \\
\degname{semantic tool tissue interaction mismatch} & 1.00 & -- & -- & 0.25 & -- & -- \\
\degname{instrument jitter} & -- & -- & -- & 1.00 & -- & 0.70 \\
\degname{instrument morph} & -- & -- & -- & 1.00 & -- & 0.55 \\
\degname{instrument partial disappearance} & -- & -- & -- & 1.00 & -- & 0.65 \\
\degname{temporally inconsistent instrument duplication} & -- & -- & -- & 1.00 & -- & 0.60 \\
\degname{tissue deformation} & -- & -- & -- & -- & 1.00 & 0.55 \\
\degname{physical tool tissue response inconsistency} & -- & -- & -- & -- & 1.00 & 0.35 \\
\degname{tissue boundary melting or splitting} & -- & -- & -- & -- & 1.00 & 0.55 \\
\degname{nonphysical tissue pulsation} & -- & -- & -- & -- & 1.00 & 0.55 \\
\bottomrule
\end{tabular}
\end{table*}

\section{Discussion on Limitations and Broader Impact}
\label{sec:supp_limitations}

\paragraph{Limited trajectory-level action control}
FLAIR is designed as a reference-free framework that infers a plausible, text-conditioned motion prior rather than enforcing a single, user-specified action trajectory. Since a surgical prompt can admit multiple clinically valid motion realizations (e.g., varying grasp angles or interaction durations for the same instrument--action--target triplet), our text-to-action predictor is explicitly formulated to output a distributional prior \(\hat{Z}\) that captures the canonical execution pattern learned from real surgical videos, rather than committing to an arbitrary frame-level trajectory (Eq.~9 in main paper). This design choice is intentional: by constraining generation to the compact, action-conditioned region observed in real data, FLAIR already substantially narrows the space of plausible outputs compared to unconstrained text-to-video generation, as reflected in its improved motion stability and action consistency (Table~1, Table~2 in main paper). Nonetheless, offering more explicit, trajectory- or keyframe-level steering on top of this learned prior, while preserving the reference-free inference setting, remains a natural extension for future work.

\paragraph{Coverage constrained by the BSA-10 action taxonomy and source datasets.}
SurgActionClip-30K is built by consolidating 11 diverse surgical datasets under the cross-specialty BSA-10 taxonomy~\citep{xu2026generalized}, which was deliberately chosen for its broad coverage across laparoscopic, endoscopic, and microsurgical procedures and its validated use in cross-specialty action and skill assessment. This already provides substantially wider procedural and anatomical coverage than prior action-centric surgical datasets, spanning 6 core procedures plus 2 held-out procedures used explicitly to test generalization beyond the training taxonomy (Table~1). As with any taxonomy-driven dataset construction, motion priors learned within this scope may transfer less directly to action types or surgical modalities not represented in the source data; extending coverage to additional taxonomies and modalities is a straightforward and complementary direction, facilitated by FLAIR's modular design in which new action priors can be learned without modifying the frozen video backbone.

\paragraph{Broader impact.}
FLAIR and SurgActionClip-30K are intended to support surgical education, simulation, and data augmentation in settings where real annotated surgical video is scarce or costly to obtain~\citep{eckhoff2023sages,king2025use}, while SurgMetrics aims to give the community a standardized, surgery-aware way to evaluate progress in this emerging area. Together, these three contributions point toward a future in which surgical video generation can scale beyond small, curated datasets and manually inspected results, lowering the barrier for institutions with limited resources to build simulation and training pipelines. We emphasize that FLAIR-generated videos and their SurgMetrics scores are intended as research and educational tools rather than validated clinical resources, and any real-world deployment should include expert oversight and clear labeling of synthetic content.

\end{document}

%% file: math_commands.tex
\usepackage{amsmath,amsfonts,bm}

\def\eqref#1{equation~\ref{#1}}

\def\1{\bm{1}}

\DeclareMathAlphabet{\mathsfit}{\encodingdefault}{\sfdefault}{m}{sl}
\SetMathAlphabet{\mathsfit}{bold}{\encodingdefault}{\sfdefault}{bx}{n}

